\documentclass[]{article}
\usepackage{hyperref}

\usepackage[font=small]{caption}

\usepackage{amsfonts}
\usepackage{natbib}
\usepackage{breakcites}
\usepackage{authblk}
\usepackage{amsmath}
\usepackage{graphicx}
\usepackage{bm}
\usepackage{listings}

\newif\ifMIT
\MITfalse

\newcommand{\newterm}[1]{{\bf #1}}

\def\figref#1{figure~\ref{#1}}
\def\Figref#1{Figure~\ref{#1}}

\def\secref#1{section~\ref{#1}}
\def\eqref#1{equation~\ref{#1}}
\def\1{\bm{1}}

\def\vx{{\bm{x}}}

\def\vz{{\bm{z}}}
\ifMIT
\def\gamma{{\boldmath{\vgamma}}}
\def\vtheta{{\boldmath{\theta}}}

\else

\def\vgamma{{\bm{\gamma}}}
\def\vtheta{{\bm{\theta}}}

\fi

\def\evx{{x}}

\ifMIT

\else

\fi

\DeclareMathAlphabet{\mathsfit}{\encodingdefault}{\sfdefault}{m}{sl}
\SetMathAlphabet{\mathsfit}{bold}{\encodingdefault}{\sfdefault}{bx}{n}

\newcommand{\pdata}{p_{\rm{data}}}
\newcommand{\ptrain}{\hat{p}_{\rm{data}}}
\newcommand{\pmodel}{p_{\rm{model}}}

\newcommand{\E}{\mathbb{E}}

\newcommand{\KL}{D_{\mathrm{KL}}}

\ifMIT
\else
\DeclareMathOperator*{\argmax}{arg\,max}
\DeclareMathOperator*{\argmin}{arg\,min}
\fi

\title{NIPS 2016 Tutorial:\\ Generative Adversarial Networks}

\author{Ian Goodfellow}
\affil{OpenAI, {\tt ian@openai.com}}

\date{}

\DeclareRobustCommand{\[}{\begin{equation}}
\DeclareRobustCommand{\]}{\end{equation}}

\begin{document}

\newlength{\figwidth}
\setlength{\figwidth}{26pc}

\maketitle

\begin{abstract}
This report summarizes the tutorial presented by the author at NIPS 2016 on
{\em generative adversarial networks} (GANs).
The tutorial describes:
(1) Why generative modeling is a topic worth studying,
(2) how generative models work, and how GANs compare to other generative models,
(3) the details of how GANs work,
(4) research frontiers in GANs, and
(5) state-of-the-art image models that combine GANs with other methods.
Finally, the tutorial contains three exercises for readers to complete,
and the solutions to these exercises.
\end{abstract}

\section*{Introduction}

This report\footnote{This is the arxiv.org version of this tutorial.
  Some graphics have been compressed to respect arxiv.org's 10MB limit on paper size,
  and do not reflect the full image quality.
}
summarizes the content of the NIPS 2016 tutorial on {\em generative adversarial networks}
(GANs) \citep{Goodfellow-et-al-NIPS2014-small}.
The tutorial was designed primarily to ensure that it answered most of
the questions asked by audience members ahead of time, in order to make sure
that the tutorial would be as useful as possible to the audience.
This tutorial is not intended to be a comprehensive review of the field
of GANs; many excellent papers are not described here, simply because
they were not relevant to answering the most frequent questions, and because
the tutorial was delivered as a two hour oral presentation and did not
have unlimited time cover all subjects.

The tutorial describes:
(1) Why generative modeling is a topic worth studying,
(2) how generative models work, and how GANs compare to other generative models,
(3) the details of how GANs work,
(4) research frontiers in GANs, and
(5) state-of-the-art image models that combine GANs with other methods.
Finally, the tutorial contains three exercises for readers to complete,
and the solutions to these exercises.

The slides for the tutorial are available in PDF and Keynote format at the following URLs:

\url{http://www.iangoodfellow.com/slides/2016-12-04-NIPS.pdf}

\url{http://www.iangoodfellow.com/slides/2016-12-04-NIPS.key}

The video was recorded by the NIPS foundation and should be made available at a later date.

\begin{figure}
\center
\includegraphics[width=\textwidth]{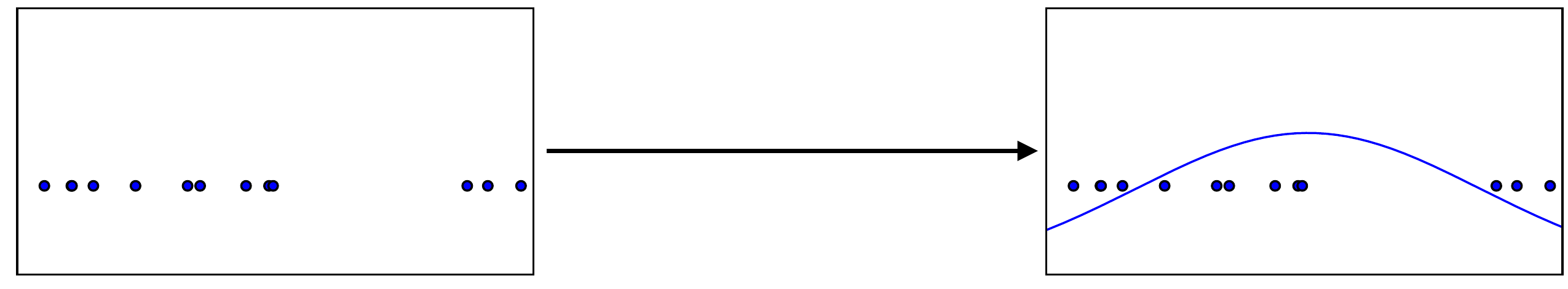}
\caption{Some generative models perform density estimation.
These models take a training set of examples drawn from an unknown
data-generating distribution $\pdata$ and return an estimate of that
distribution. The estimate $\pmodel$ can be evaluated for a particular
value of $\vx$ to obtain an estimate $\pmodel(\vx)$ of the true
density $\pmodel(\vx)$.
This figure illustrates the process for a collection of samples of
one-dimensional data and a Gaussian model.
}
\label{fig:density}
\end{figure}

\begin{figure}
  \center
  \includegraphics[width=\textwidth]{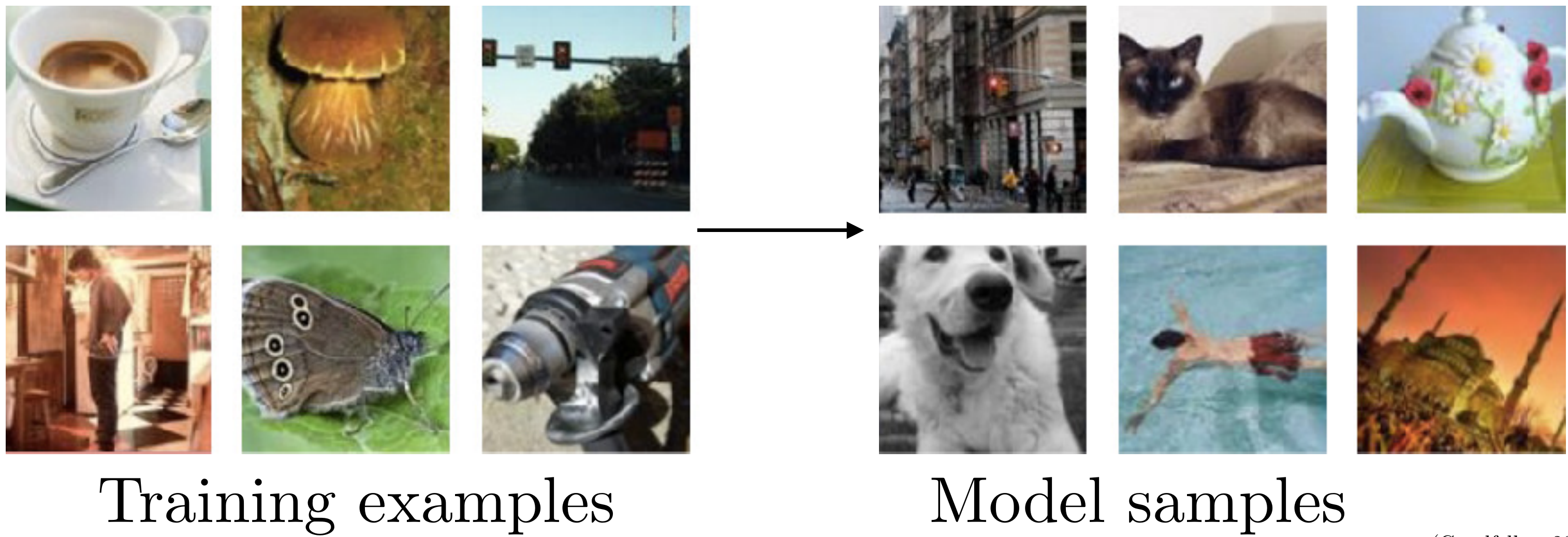}
  \caption{Some generative models are able to generate samples
    from the model distribution.
    In this illustration of the process, we show samples from
    the ImageNet \citep{imagenet_cvpr09,Deng2010,ILSVRCarxiv14} 
    dataset.
    An ideal generative model would be able to train on examples
    as shown on the left and then create more examples from the
    same distribution as shown on the right.
    At present, generative models are not yet advanced enough to
    do this correctly for ImageNet, so for demonstration purposes
    this figure uses actual ImageNet data to illustrate what an
    ideal generative model would produce.
  }
  \label{fig:generative_machine}
\end{figure}

Generative adversarial networks are an example of {\em generative models}.
The term ``generative model'' is used in many different ways.
In this tutorial, the term refers to any model that takes a training set,
consisting of samples drawn from a distribution $\pdata$, and learns to
represent an estimate of that distribution somehow.
The result is a probability distribution $\pmodel$.
In some cases, the model estimates $\pmodel$ explicitly, as shown in
\figref{fig:density}.
In other cases, the model is only able to generate samples from
$\pmodel$, as shown in \figref{fig:generative_machine}.
Some models are able to do both.
GANs focus primarily on sample generation, though it is possible to
design GANs that can do both.

\section{Why study generative modeling?}
\label{sec:why}

One might legitimately wonder why generative models are worth studying,
especially generative models that are only capable of generating data
rather than providing an estimate of the density function.
After all, when applied to images, such models seem to merely provide
more images, and the world has no shortage of images.

\begin{figure}
\centering
\includegraphics[width=\textwidth]{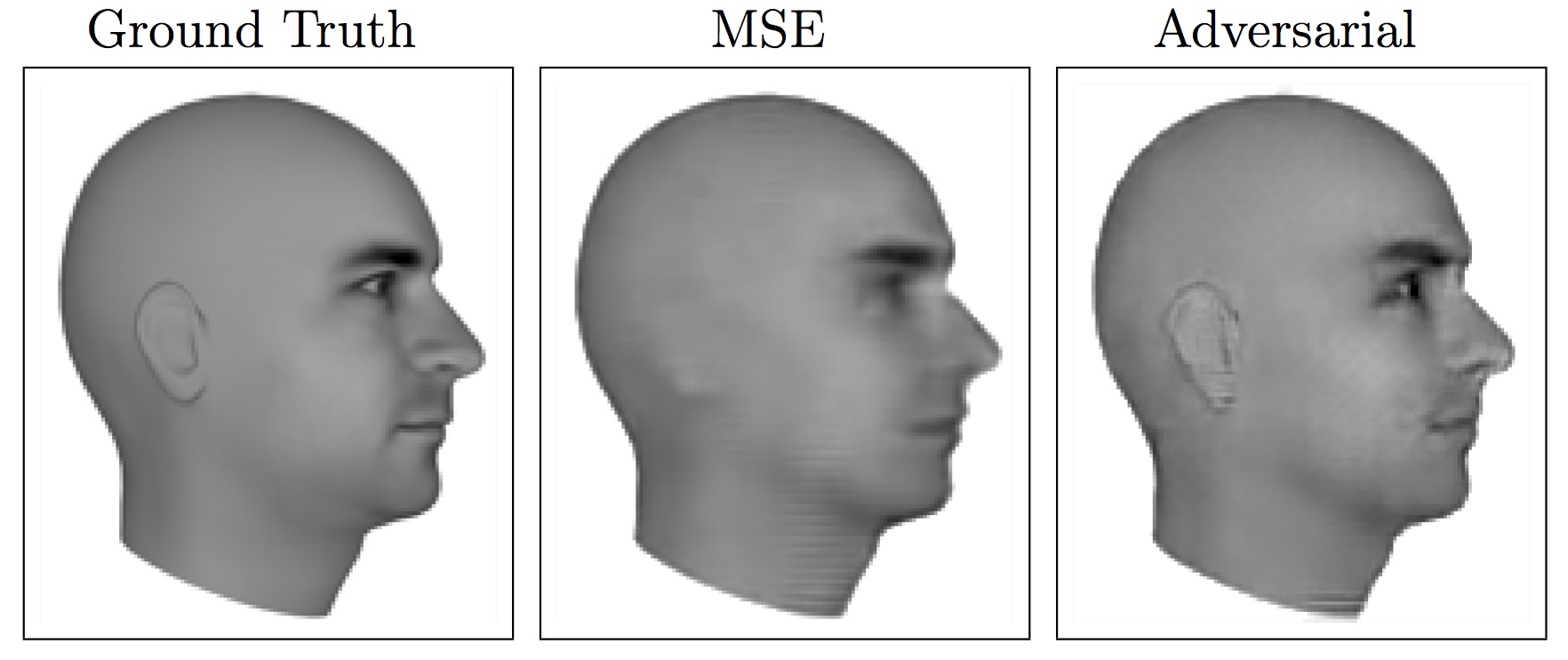}
\caption{
\citet{lotter2015unsupervised} provide an excellent illustration of the importance
of being able to model multi-modal data.
In this example, a model is trained to predict the next frame in a video sequence.
The video depicts a computer rendering of a moving 3D model of a person's head.
The image on the left shows an example of an actual frame of video, which the model
would ideally predict.
The image in the center shows what happens when the model is trained using mean
squared error between the actual next frame and the model's predicted next frame.
The model is forced to choose a single answer for what the next frame will look like.
Because there are many possible futures, corresponding to slightly different positions
of the head, the single answer that the model chooses corresponds to an average over
many slightly different images.
This causes the ears to practically vanish and the eyes to become blurry.
Using an additional GAN loss, the image on the right is able to understand that there
are many possible outputs, each of which is sharp and recognizable as a realistic,
detailed image.
}
  \label{fig:lotter}
\end{figure}

There are several reasons to study generative models, including:
\begin{itemize}

\item Training and sampling from generative models is an excellent test
of our ability to represent and manipulate high-dimensional probability
distributions.
High-dimensional probability distributions are important objects in a
wide variety of applied math and engineering domains.

\item Generative models can be incorporated into reinforcement learning in several
  ways.
  Reinforcement learning algorithms can be divided into two categories;
  model-based and model-free, with model-based algorithms being those that
  contain a generative model.
  Generative models of time-series data can be used to simulate possible
futures. Such models could be used for planning and for reinforcement learning
in a variety of ways.
A generative model used for planning can learn a conditional distribution over
future states of the world, given the current state of the world and hypothetical
actions an agent might take as input.
The agent can query the model with different potential actions and choose actions
that the model predicts are likely to yield a desired state of the world.
For a recent example of such a model, see \citet{finn2016unsupervised},
and for a recent example of the use of such a model for planning,
see \citet{finn2016deep}. 
Another way that generative models might be used for reinforcement learning is
to enable learning in an imaginary environment, where mistaken actions do not
cause real damage to the agent.
Generative models can also be used to guide exploration by keeping track of
how often different states have been visited or different actions have been
attempted previously.
Generative models, and especially GANs, can also be used for inverse reinforcement
learning.
Some of these connections to reinforcement learning are described further in
\secref{sec:rl_connections}.

\item Generative models can be trained with missing data and can provide predictions
  on inputs that are missing data.
  One particularly interesting case of missing data is {\em semi-supervised learning},
  in which the labels for many or even most training examples are missing.
  Modern deep learning algorithms typically require extremely many labeled examples
  to be able to generalize well.
  Semi-supervised learning is one strategy for reducing the number of labels.
  The learning algorithm can improve its generalization by studying a large number
  of unlabeled examples which, which are usually easier to obtain.
  Generative models, and GANs in particular, are able to perform semi-supervised
  learning reasonably well. This is described further in \secref{sec:ssl}.

\item Generative models, and GANs in particular, enable machine learning to work with
  {\em multi-modal} outputs.
  For many tasks, a single input may correspond to many different correct answers,
  each of which is acceptable.
Some traditional means of training machine learning models, such as minimizing the
mean squared error between a desired output and the model's predicted output, are
not able to train models that can produce multiple different correct answers.
One example of such a scenario is predicting the next frame in a video, as shown
in \figref{fig:lotter}.

\item Finally, many tasks intrinsically require realitic generation of samples from
  some distribution.
\end{itemize}

\begin{figure}
  \centering
  \includegraphics[width=\textwidth]{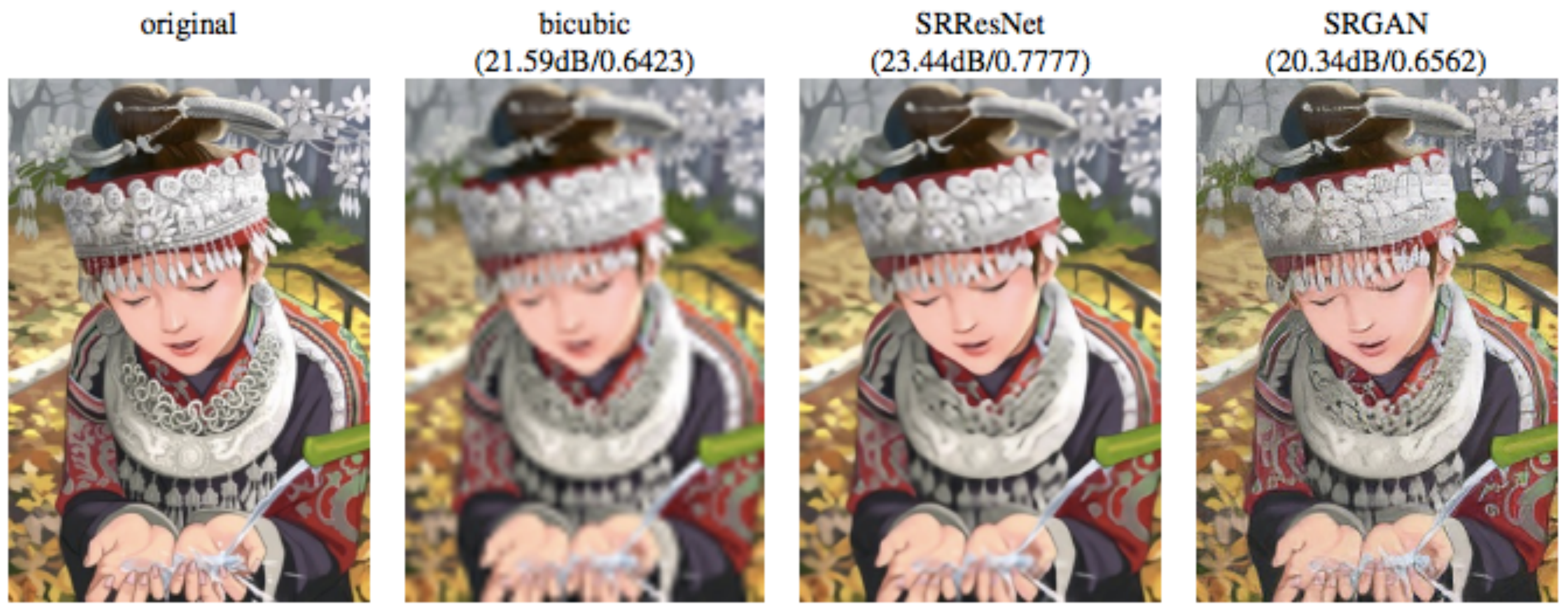}
  \caption{
\citet{Ledig16} demonstrate excellent single-image superresolution results that show
the benefit of using a generative model trained to generate realistic samples from
 a multimodal distribution.
 The leftmost image is an original high-resolution image.
 It is then downsampled to make a low-resolution image, and different methods
 are used to attempt to recover the high-resolution image.
 The bicubic method is simply an interpolation method that does not use 
 the statistics of the training set at all.
 SRResNet is a neural network trained with mean squared error.
 SRGAN is a GAN-based neural network that improves over SRGAN because it is able
 to understand that there are multiple correct answers, rather than averaging
 over many answers to impose a single best output.
  }
  \label{fig:superres}
\end{figure}

\begin{figure}
  \centering
  \includegraphics[width=\textwidth]{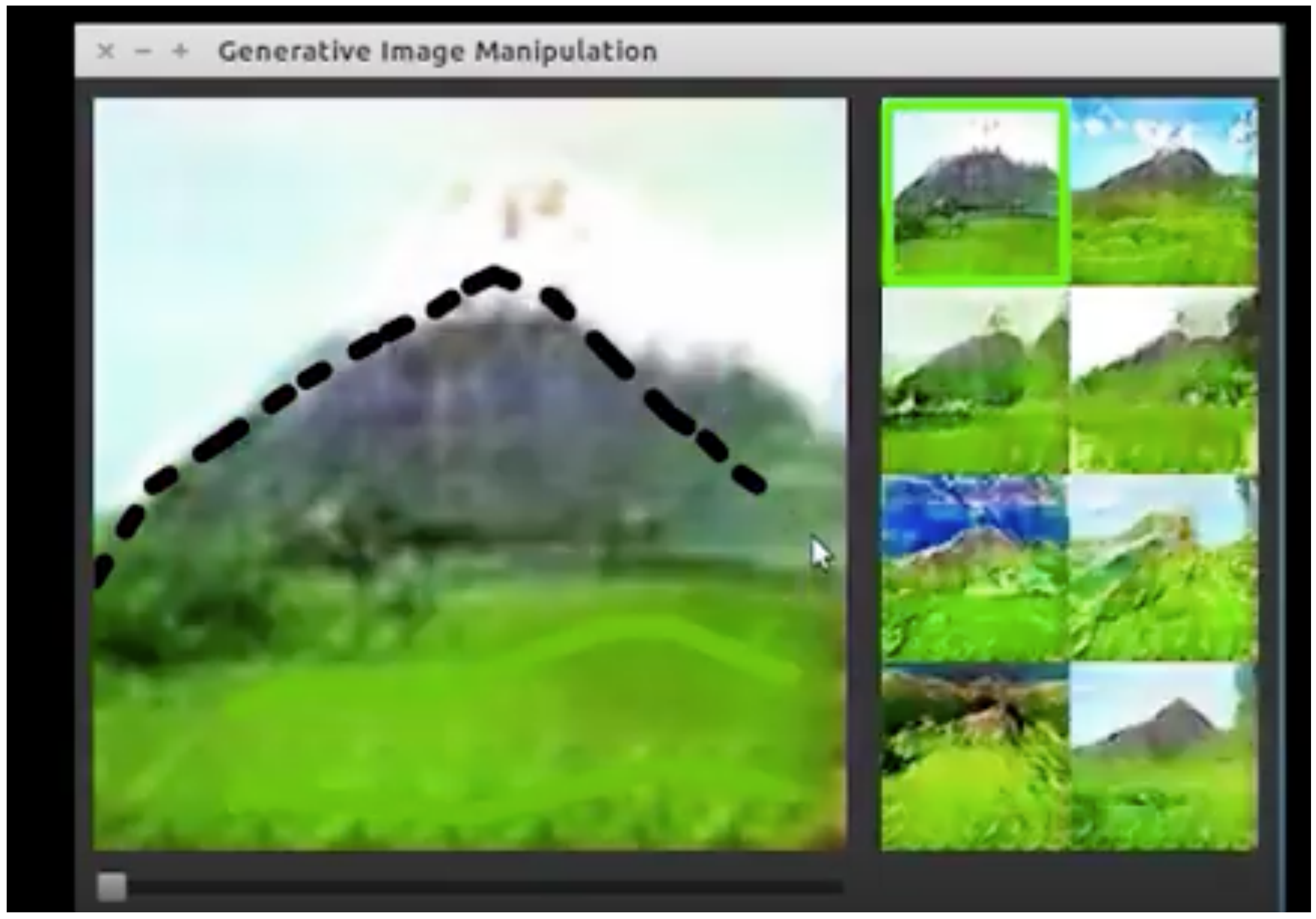}
  \caption{
    \citet{zhu2016generative} developed an interactive application called {\em interactive generative adversarial networks}
(iGAN). 
A user can draw a rough sketch of an image, and iGAN uses a GAN to produce the most similar
realistic image.
In this example, a user has scribbled a few green lines that iGAN has converted into a grassy
field, and the user has drawn a black triangle that iGAN has turned into a detailed mountain.
Applications that create art are one of many reasons to study generative models that create
images.
A video demonstration of iGAN is available at the following URL:
\url{https://www.youtube.com/watch?v=9c4z6YsBGQ0}
}
  \label{fig:igan}
\end{figure}

\begin{figure}
  \centering
  \includegraphics[width=\textwidth]{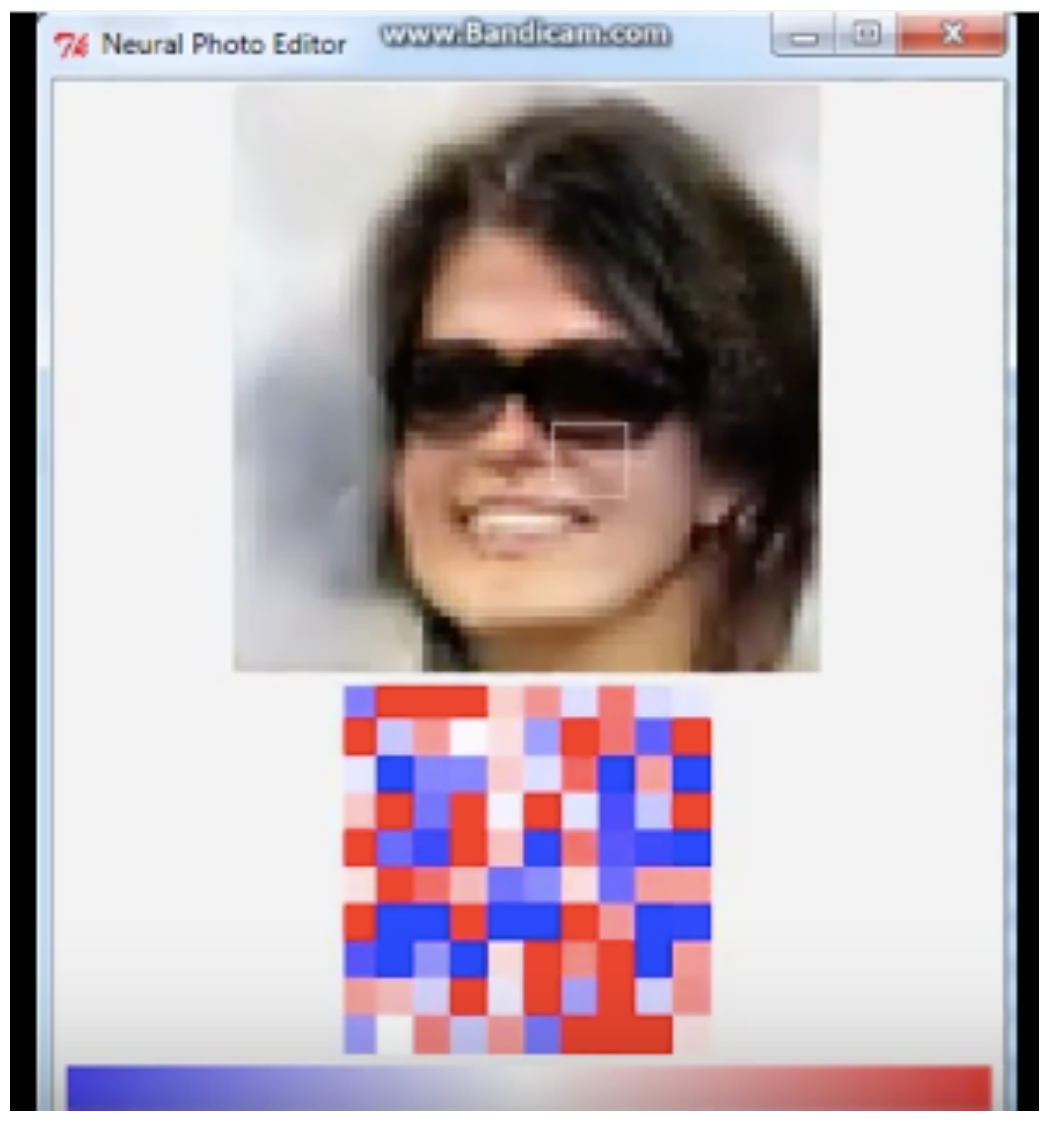}
  \caption{
    \citet{BrockLRW16a} developed {\em introspective adversarial networks} (IAN).
    The user paints rough modifications to a photo, such as painting
    with black paint in an area where the user would like to add black
    hair, and IAN turns these rough paint strokes into photorealistic
    imagery matching the user's desires.
    Applications that enable a user to make realistic modifications to
    photo media are one of many reasons to study generative models
    that create images.
    A video demonstration of IAN is available at the following URL:
    \url{https://www.youtube.com/watch?v=FDELBFSeqQs}
  }
  \label{fig:ian}
\end{figure}

\begin{figure}
  \centering
  \includegraphics[width=\textwidth]{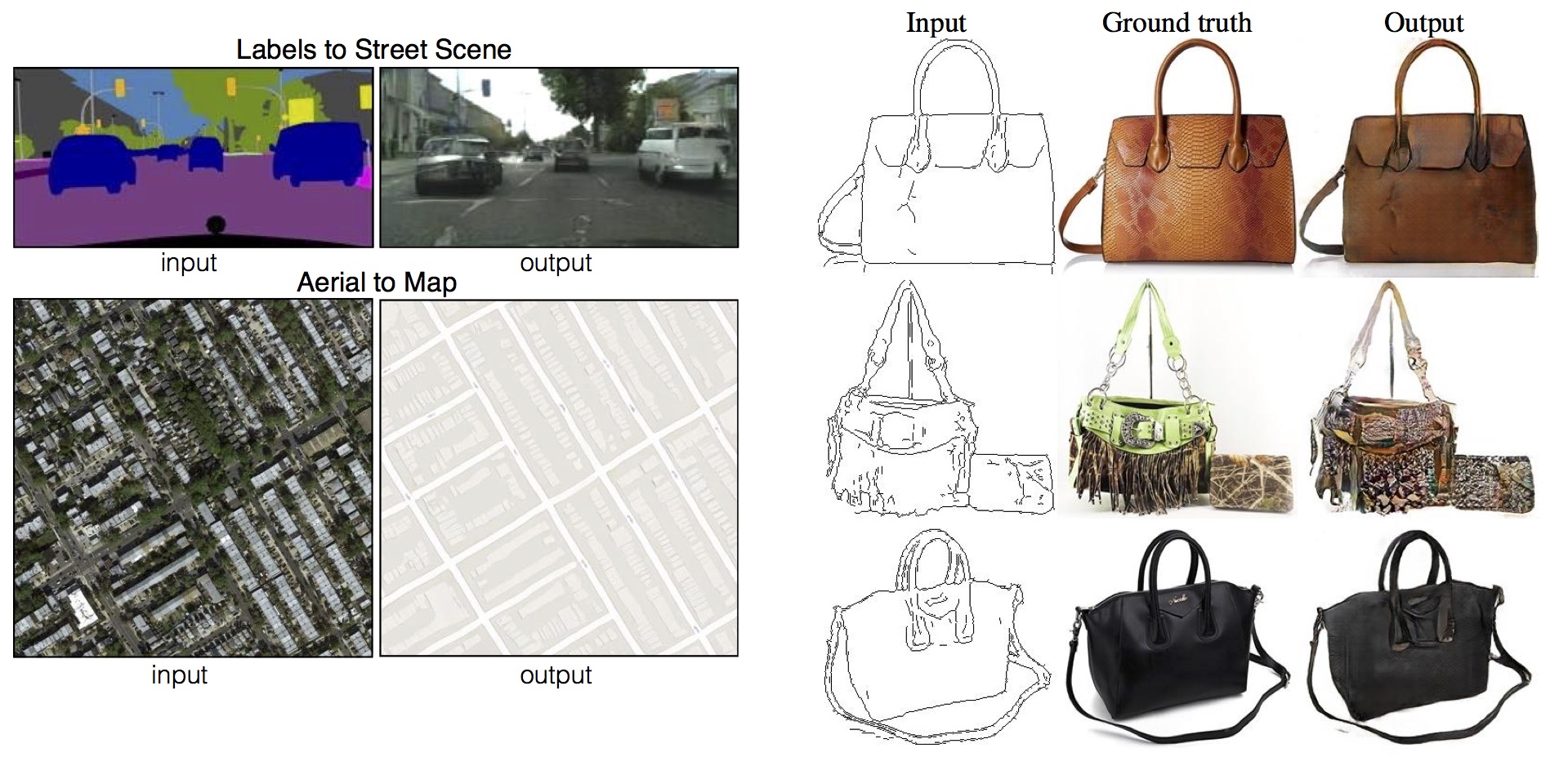}
  \caption{
    \citet{isola2016image}
    created a concept they called {image to image translation},
    encompassing many kinds of transformations of an image:
    converting a satellite photo into a map,
    coverting a sketch into a photorealistic image,
    etc.
    Because many of these conversion processes have multiple
    correct outputs for each input, it is necessary to use
    generative modeling to train the model correctly.
    In particular, \citet{isola2016image} use a GAN.
    Image to image translation provides many examples of how
    a creative algorithm designer can find several unanticipated uses
    for generative models.
    In the future, presumably many more such creative uses
    will be found.
  }
  \label{fig:im2im}
\end{figure}

Examples of some of these tasks that intrinsically require the generation of good
samples include:
\begin{itemize}
  \item {\em Single image super-resolution}: In this task, the goal is to take a
    low-resolution image and synthesize a high-resolution equivalent.
    Generative modeling is required because this task requires the model to impute
    more information into the image than was originally there in the input.
    There are many possible high-resolution images corresponding to the low-resolution
    image.
    The model should choose an image that is a sample from the probability distribution
    over possible images.
    Choosing an image that is the average of all possible images would yield a result
    that is too blurry to be pleasing.
    See \figref{fig:superres}.

  \item Tasks where the goal is to create art.
    Two recent projects have both demonstrated that generative models, and in particular,
    GANs, can be used to create interactive programs that assist the user in creating
    realistic images that correspond to rough scenes in the user's imagination.
    See \figref{fig:igan} and \figref{fig:ian}.

  \item Image-to-image translation applications can convert aerial photos into maps
    or convert sketches to images. There is a very long tail of creative applications
    that are difficult to anticipate but useful once they have been discovered.
    See \figref{fig:im2im}.

\end{itemize}

All of these and other applications of generative models provide compelling
reasons to invest time and resources into improving generative models.

\section{How do generative models work? How do GANs compare to others?}
\label{sec:tree}

We now have some idea of what generative models can do and why it might be
desirable to build one.
Now we can ask: how does a generative model actually work? And in particular,
how does a GAN work, in comparison to other generative models?

\subsection{Maximum likelihood estimation}

To simplify the discussion somewhat, we will focus on generative models
that work via the principle of \newterm{maximum likelihood}.
Not every generative model uses maximum likelihood.
Some generative models do not use maximum likelihood by default, but
can be made to do so (GANs fall into this category).
By ignoring those models that do not use maximum likelihood, and
by focusing on the maximum likelihood version of models that do not
usually use maximum likelihood, we can eliminate some of the more 
distracting differences between different models.

The basic idea of maximum likelihood is to define a model that provides
an estimate of a probability distribution, parameterized by parameters
$\vtheta$.
We then refer to the \newterm{likelihood} as the probability that the model
assigns to the training data: $\prod_{i=1}^m \pmodel\left(\vx^{(i)}; \vtheta \right),$
for a dataset containing $m$ training examples $\vx^{(i)}$.

The principle of maximum likelihood simply says to choose the parameters for the model
that maximize the likelihood of the training data.
This is easiest to do in log space, where we have a sum rather than a product
over examples.
This sum simplifies the algebraic expressions for the derivatives of the likelihood
with respect to the models, and when implemented on a digital computer, is less
prone to numerical problems, such as underflow resulting from multiplying together
several very small probabilities.

\begin{align}
\vtheta^* =& \argmax_\vtheta \prod_{i=1}^m \pmodel\left(\vx^{(i)}; \vtheta \right) \\
  =& \argmax_\vtheta \log \prod_{i=1}^m \pmodel\left(\vx^{(i)}; \vtheta \right) \label{eq:log} \\
          =& \argmax_\vtheta \sum_{i=1}^m \log \pmodel\left(\vx^{(i)}; \vtheta \right).
\end{align}

In \eqref{eq:log}, we have used the property that $\argmax_v f(v) = \argmax_v \log f(v)$ for 
positive $v$, because the logarithm is a function that increases everywhere and does not change
the location of the maximum.

\begin{figure}
\centering
\includegraphics[width=\textwidth]{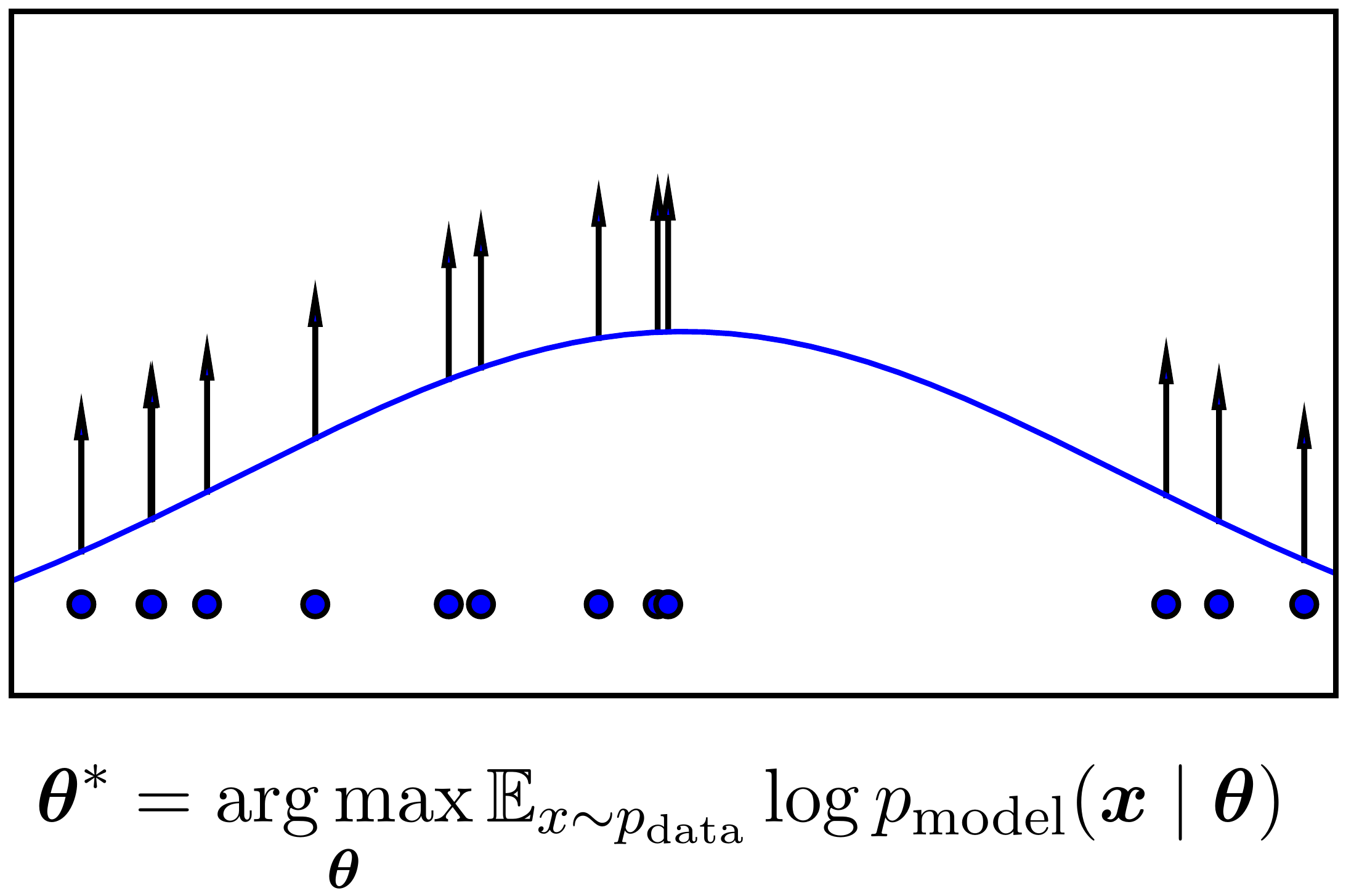}
\caption{The maximum likelihood process consists of taking several samples from
  the data generating distribution to form a training set, then pushing up on the
  probability the model assigns to those points, in order to maximize the likelihood
  of the training data.
  This illustration shows how different data points push up on different parts of
  the density function for a Gaussian model applied to 1-D data.
  The fact that the density function must sum to $1$ means that we cannot simply
  assign infinite likelihood to all points; as one point pushes up in one place
  it inevitably pulls down in other places.
  The resulting density function balances out the upward forces from all the data
  points in different locations.
}
\label{fig:mle}
\end{figure}

The maximum likelihood process is illustrated in \figref{fig:mle}.

We can also think of maximum likelihood estimation as minimizing the
\newterm{KL divergence} between the data generating distribution and the
model:
\begin{equation}
\vtheta^* = \argmin_\vtheta \KL\left( \pdata(\vx) \Vert \pmodel(\vx ; \vtheta) \right).
\label{eq:kl}
\end{equation}
If we were able to do this precisely, then if $\pdata$ lies within the family of distributions
$\pmodel(\vx ; \vtheta)$, the model would recover $\pdata$ exactly.
In practice, we do not have access to $\pdata$ itself, but only to a training set
consisting of $m$ samples from $\pdata$.
We uses these to define $\ptrain$, an \newterm{empirical distribution} that places mass only
on exactly those $m$ points, approximating $\pdata$.
Minimizing the KL divergence between $\ptrain$ and $\pmodel$ is exactly equivalent to maximizing
the log-likelihood of the training set.

For more information on maximum likelihood and other statistical estimators,
see chapter 5 of \citet{Goodfellow-et-al-2016}.

\subsection{A taxonomy of deep generative models}

\begin{figure}
\centering
\includegraphics[width=\textwidth]{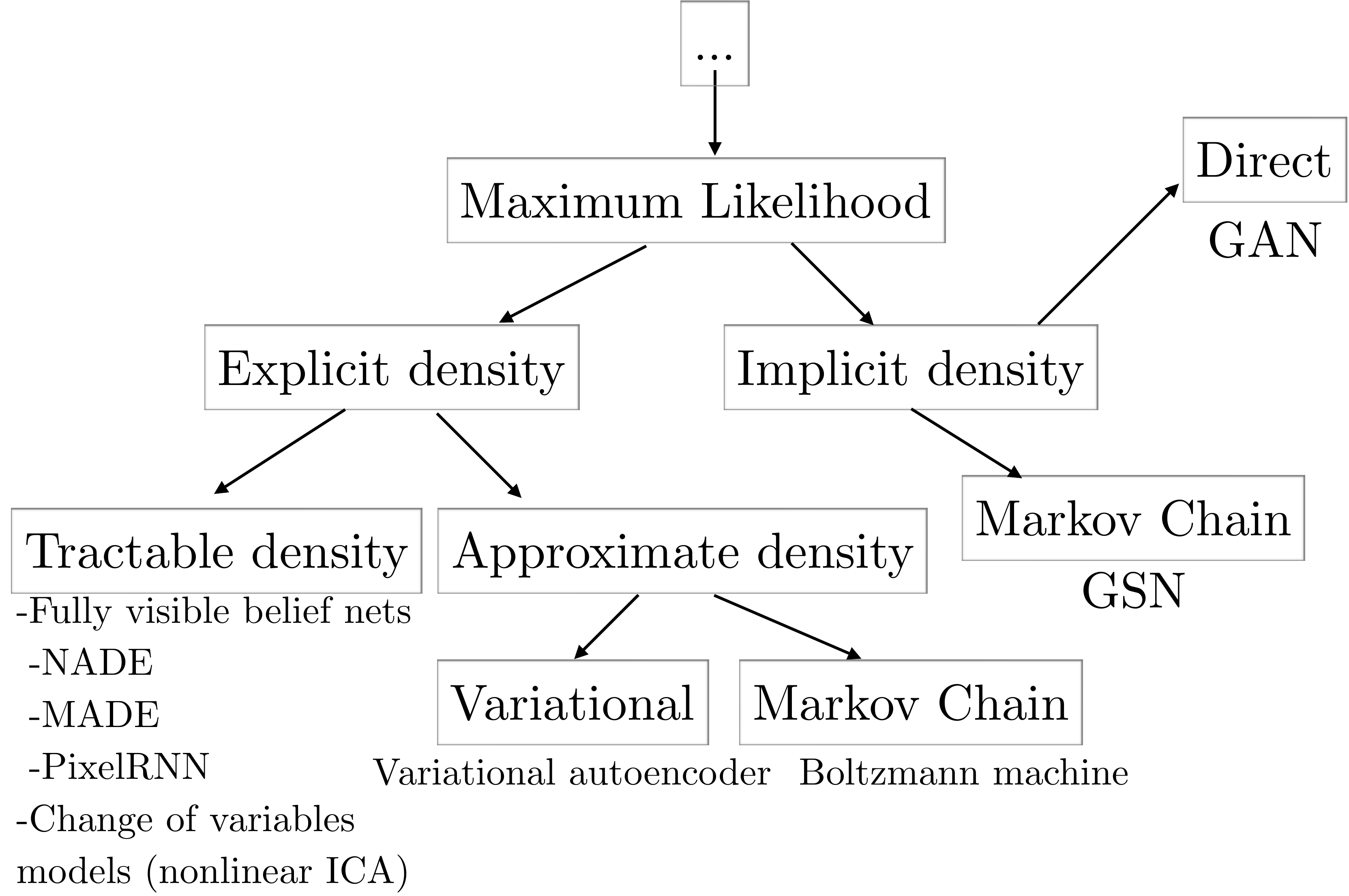}
\caption{
Deep generative models that can learn via the principle of maximim likelihood
differ with respect to how they represent or approximate the likelihood.
On the left branch of this taxonomic tree, models construct an explicit density,
$\pmodel(\vx; \vtheta)$, and thus an explicit likelihood which can be maximized.
Among these explicit density models, the density may be computationally tractable,
or it may be intractable, meaning that to maximize the likelihood it is necessary
to make either variatioanl approximations or Monte
Carlo approximations (or both).
On the right branch of the tree, the model does not explicitly represent a
probability distribution over the space where the data lies.
Instead, the model provides some way of interacting less directly with this
probability distribution.
Typically the indirect means of interacting with the probability distribution is
the ability to draw samples from it.
Some of these implicit models that offer the ability to sample from the distribution
do so using a Markov Chain; the model defines a way to stochastically transform
an existing sample in order to obtain another sample from the same distribution.
Others are able to generate a sample in a single step, starting without any input.
While the models used for GANs can sometimes be constructed to define an explicit
density, the training algorithm for GANs makes use only of the model's ability to
generate samples.
GANs are thus trained using the strategy from the rightmost leaf of the tree:
using an implicit model that samples directly from the distribution represented
by the model.
}
\label{fig:tree}
\end{figure}

If we restrict our attention to deep generative models that work by maximizing
the likelihood, we can compare several models by contrasting the ways that they
compute either the likelihood and its gradients, or approximations to these
quantities.
As mentioned earlier, many of these models are often used with principles other
than maximum likelihood, but we can examine the maximum likelihood variant of
each of them in order to reduce the amount of distracting differences between
the methods.
Following this approach, we construct the taxonomy shown in \figref{fig:tree}.
Every leaf in this taxonomic tree has some advantages and disadvantages.
GANs were designed to avoid many of the disadvantages present in pre-existing
nodes of the tree, but also introduced some new disadvantages.

\subsection{Explicit density models}

In the left branch of the taxonomy shown in \figref{fig:tree} are models that define
an explicit density function $\pmodel(\vx ; \vtheta)$.
For these models, maxmimization of the likelihood is straightforward; we simply plug
the model's definition of the density function into the expression for the likelihood,
and follow the gradient uphill.

The main difficulty present in explicit density models is designing a model that can
capture all of the complexity of the data to be generated 
while still maintaining computational tractability.
There are two different strategies used to confront this challenge:
(1) careful construction of models whose structure guarantees their tractability,
as described in \secref{sec:explicit_tractable},
and (2) models that admit tractable approximations to the likelihood and its
gradients, as described in \secref{sec:approx}.

\subsubsection{Tractable explicit models}
\label{sec:explicit_tractable}

In the leftmost leaf of the taxonomic tree of \figref{fig:tree} are the models
that define an explicit density function that is computationally tractable.
There are currently two popular approaches to tractable explicit density models:
fully visible belief networks and nonlinear independent components analysis.

\paragraph{Fully visible belief networks}
\newterm{Fully visible belief networks} \citep{Frey96,Frey98} or FVBNs are models that use the chain
rule of probability to decompose a probability distribution over an $n$-dimensional vector $\vx$
into a product of one-dimensional probability distributions:
\[
\pmodel(\vx) = \prod_{i=1}^n \pmodel\left(\evx_i \mid \evx_1, \dots, \evx_{i-1} \right).
\]
FVBNs are, as of this writing, one of the three most popular
approaches to generative modeling, alongside GANs and variational autoencoders.
They form the basis for sophisticated generative models from DeepMind, such as
WaveNet \citep{aaron-wavenet-2016}. WaveNet is able to generate realistic human speech.
The main drawback of FVBNs is that samples must be generated
one entry at a time: first $\evx_1$, then $\evx_2$, etc., so the cost of generating
a sample is $O(n)$.
In modern FVBNs such as WaveNet, the distribution over each $\evx_i$ is computed by a deep
neural network, so each of these $n$ steps involves a nontrivial amount of computation.
Moreover, these steps cannot be parallelized.
WaveNet thus requires two minutes of computation time to generate one second of audio,
and cannot yet be used for interactive conversations.
GANs were designed to be able to generate all of $\vx$ in parallel, yielding greater
generation speed.

\begin{figure}
\centering
\includegraphics[width=\textwidth]{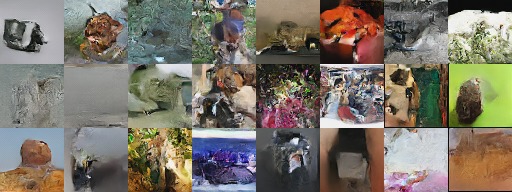}
\caption{Samples generated by a real NVP model trained on 64x64 ImageNet images.
Figure reproduced from \citet{dinh2016density}.}
\label{fig:nvp}
\end{figure}

\paragraph{Nonlinear independent components analysis}
Another family of deep generative models with explicit density functions is based on
defining continuous, nonlinear transformations between two different spaces.
For example, if there is a vector of latent variables $\vz$ and a continuous, differentiable,
invertible transformation
$g$ such that $g(\vz)$ yields a sample from the model in $\vx$ space,
then
\begin{equation}
  \label{eq:change-of-variable}
  p_x(\vx) = p_z(g^{-1}(\vx)) \left| \mathrm{det}
  \left( \frac{\partial g^{-1}(\vx)} {\partial \vx}\right) \right|. 
\end{equation}
The density $p_x$ is tractable if the density $p_z$ is tractable
and the determinant of the Jacobian of $g^{-1}$ is tractable.
In other words, a simple distribution over $\vz$ combined with
a transformation $g$ that warps space in complicated ways can yield
a complicated distribution over $\vx$, and if $g$ is carefully designed,
the density is tractable too.
Models with nonlinear $g$ functions date back at least to
~\citet{deco1995higher}.
The latest member of this family is real NVP \citep{dinh2016density}.
See \figref{fig:nvp} for some visualizations of ImageNet samples
generated by real NVP.
The main drawback to nonlinear ICA models is that they impose restrictions
on the choice of the function $g$. In particular, the invertibility requirement
means that the latent variables $\vz$ must have the same dimensionality as $\vx$.
GANs were designed to impose very few requirements on $g$, and, in particular,
admit the use of $\vz$ with larger dimension than $\vx$.

For more information about the chain rule of probability used to define FVBNs
or about the effect of deterministic transformations on probability densities
as used to define nonlinear ICA models, see chapter 3 of \citet{Goodfellow-et-al-2016}.

In summary, models that define an explicit, tractable density are highly
effective, because they permit the use of an optimization algorithm directly
on the log-likelihood of the training data.
However, the family of models that provide a tractable density is limited,
with different families having different disadvantages.

\subsubsection{Explicit models requiring approximation}
\label{sec:approx}

To avoid some of the disadvantages imposed by the design requirements of models
with tractable density functions, other models have been developed that still
provide an explicit density function but use one that is intractable, requiring
the use of approximations to maximize the likelihood.
These fall roughly into two categories: those using deterministic approximations,
which almost always means variational methods, and those using stochastic approximations,
meaning Markov chain Monte Carlo methods.

\begin{figure}
  \centering
  \includegraphics[width=\textwidth]{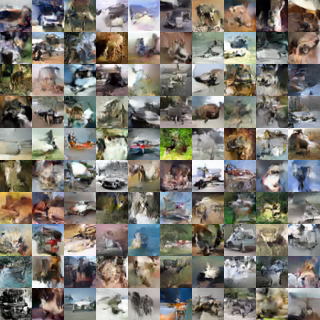}
  \caption{Samples drawn from a VAE trained on the CIFAR-10 dataset.
    Figure reproduced from \citet{kingma2016improving}.
  }
  \label{fig:vae_samples}
\end{figure}

\paragraph{Variational approximations}
Variational methods define a lower bound
\[ \mathcal{L}(\vx; \vtheta) \leq \log \pmodel(\vx; \vtheta). \]
A learning algorithm that maximizes $\mathcal{L}$ is guaranteed to obtain at least
as high a value of the log-likelihood as it does of $\mathcal{L}$.
For many families of models, it is possible to define an $\mathcal{L}$ that is computationally
tractable even when the log-likelihood is not.
Currently, the most popular approach to variational learning in deep generative models
is the \newterm{variational autoencoder} \citep{Kingma-arxiv2013,Rezende-et-al-ICML2014} or VAE.
Variational autoencoders are one of the three approaches to deep generative modeling that are
the most popular as of this writing, along with FVBNs and GANs.
The main drawback of variational methods is that,
when too weak of an approximate posterior distribution or too weak of a prior distribution is used,
\footnote{
  Empirically, VAEs with highly flexible priors or highly flexible approximate posteriors
  can obtain values of $\mathcal{L}$ that are near their own log-likelihood
  \citep{kingma2016improving,chen2016variational}.
  Of course, this is testing the gap between the objective and the bound at the maximum of the bound;
  it would be better, but not feasible, to test the gap at the maximum of the objective.
  VAEs obtain likelihoods that are competitive with other methods, suggesting that they are also
  near the maximum of the objective.
  In personal conversation, L. Dinh and D. Kingma have conjectured that a family of models
  \citep{Dinh-et-al-arxiv2014,rezende2015variational,kingma2016improving,dinh2016density}
  usable as VAE priors or approximate posteriors are universal approximators.
  If this could be proven, it would establish VAEs as being asymptotically consistent.
}
even with a perfect optimization algorithm and infinite training data, the gap
between $\mathcal{L}$ and the true likelihood can result in $\pmodel$ learning something other than
the true $\pdata$.
GANs were designed to be unbiased, in the sense that with a large enough model and infinite data,
the Nash equilibrium for a GAN game corresponds to recovering $\pdata$ exactly.
In practice, variational methods often obtain very good likelihood, but are regarded as producing
lower quality samples.
There is not a good method of quantitatively measuring sample quality, so this is a subjective opinion,
not an empirical fact.
See \figref{fig:vae_samples} for an example of some samples drawn from a VAE.
While it is difficult to point to a single aspect of GAN design and say that it results in
better sample quality, GANs are generally regarded as producing better samples.
Compared to FVBNs, VAEs are regarded as more difficult to optimize, but GANs are not
an improvement in this respect.
For more information about variational approximations, see chapter 19 of
\citet{Goodfellow-et-al-2016}.

\paragraph{Markov chain approximations}
Most deep learning algorithms make use of some form of stochastic approximation,
at the very least in the form of using a small number of randomly selected training
examples to form a minibatch used to minimize the expected loss.
Usually, sampling-based approximations work reasonably well as long as a fair sample
can be generated quickly (e.g. selecting a single example from the training set
is a cheap operation) and as long as the variance across samples is not too high.
Some models require the generation of more expensive samples, using Markov chain
techniques.
A Markov chain is a process for generating samples by repeatedly drawing a sample
$\vx' \sim q(\vx' \mid \vx).$
By repeatedly updating $\vx$ according to the transition operator $q$, Markov chain
methods can sometimes guarantee that $\vx$ will eventually converge to a sample from
$\pmodel(\vx)$.
Unfortunately, this convergence can be very slow, and there is no clear way to test
whether the chain has converged, so in practice one often uses $\vx$ too early, before
it has truly converged to be a fair sample from $\pmodel$.
In high-dimensional spaces, Markov chains become less efficient.
Boltzmann machines \citep{Fahlman83,Ackley85,Hinton-Boltzmann,Hinton86a} are a
family of generative models that rely on Markov chains both to train the model
or to generate a sample from the model.
Boltzmann machines were an important part of the deep learning renaissance beginning
in 2006 \citep{Hinton06,hinton2007learning} but they are now used only very rarely,
presumably mostly because the underlying Markov chain approximation techniques have
not scaled to problems like ImageNet generation.
Moreover, even if Markov chain methods scaled well enough to be used for training,
the use of a Markov chain to generate samples from a trained model is undesirable
compared to single-step generation methods because the multi-step Markov chain
approach has higher computational cost.
GANs were designed to avoid using Markov chains for these reasons.
For more information about Markov chain Monte Carlo approximations, see chapter 18 of
\citet{Goodfellow-et-al-2016}.
For more information about Boltzmann machines, see chapter 20 of the same book.

Some models use both variational and Markov chain approximations.
For example, deep Boltzmann machines make use of both types of
approximation \citep{SalHinton09}.

\subsection{Implicit density models}

Some models can be trained without even needing to explicitly define a density
functions.
These models instead offer a way to train the model while interacting only
indirectly with $\pmodel$, usually by sampling from it.
These constitute the second branch, on the right side, of our taxonomy of
generative models depicted in \figref{fig:tree}.

Some of these implicit models based on drawing samples from $\pmodel$ define
a Markov chain transition operator that must be run several times to obtain
 a sample from the model.
From this family, the primary example is the \newterm{generative stochastic network}
\citep{Bengio-et-al-ICML-2014}.
As discussed in \secref{sec:approx}, Markov chains often fail to scale to high
dimensional spaces, and impose increased computational costs for using the
generative model. GANs were designed to avoid these problems.

Finally, the rightmost leaf of our taxonomic tree is the family of implicit models
that can generate a sample in a single step.
At the time of their introduction, GANs were the only notable member of this family,
but since then they have been joined by additional models based on
kernelized moment matching \citep{Li-et-al-2015,dziugaite2015training}.

\subsection{Comparing GANs to other generative models}

In summary, GANs were designed to avoid many disadvantages associated with other generative
models:
\begin{itemize}
  \item They can generate samples in parallel, instead of using runtime proportional to the
    dimensionality of $\vx$. This is an advantage relative to FVBNs.
  \item The design of the generator function has very few restrictions. This is an advantage
    relative to Boltzmann machines, for which few probability distributions admit tractable
    Markov chain sampling, and relative to nonlinear ICA, for which the generator must be
    invertible and the latent code $\vz$ must have the same dimension as the samples
    $\vx$.
  \item No Markov chains are needed. This is an advantage relative to Boltzmann machines and GSNs.
  \item No variational bound is needed, and specific model families usable within the GAN
    framework are already known to be universal approximators, so GANs are already known
    to be asymptotically consistent.
    Some VAEs are conjectured to be asymptotically consistent, but this is not yet proven.
  \item GANs are subjectively regarded as producing better samples than other methods.
\end{itemize}
At the same time, GANs have taken on a new disadvantage: training them requires finding
the Nash equilibrium of a game, which is a more difficult problem than optimizing an
objective function.

\section{How do GANs work?}

We have now seen several other generative models and explained that GANs do not
work in the same way that they do. But how do GANs themselves work?

\subsection{The GAN framework}

\begin{figure}
\includegraphics[width=\textwidth]{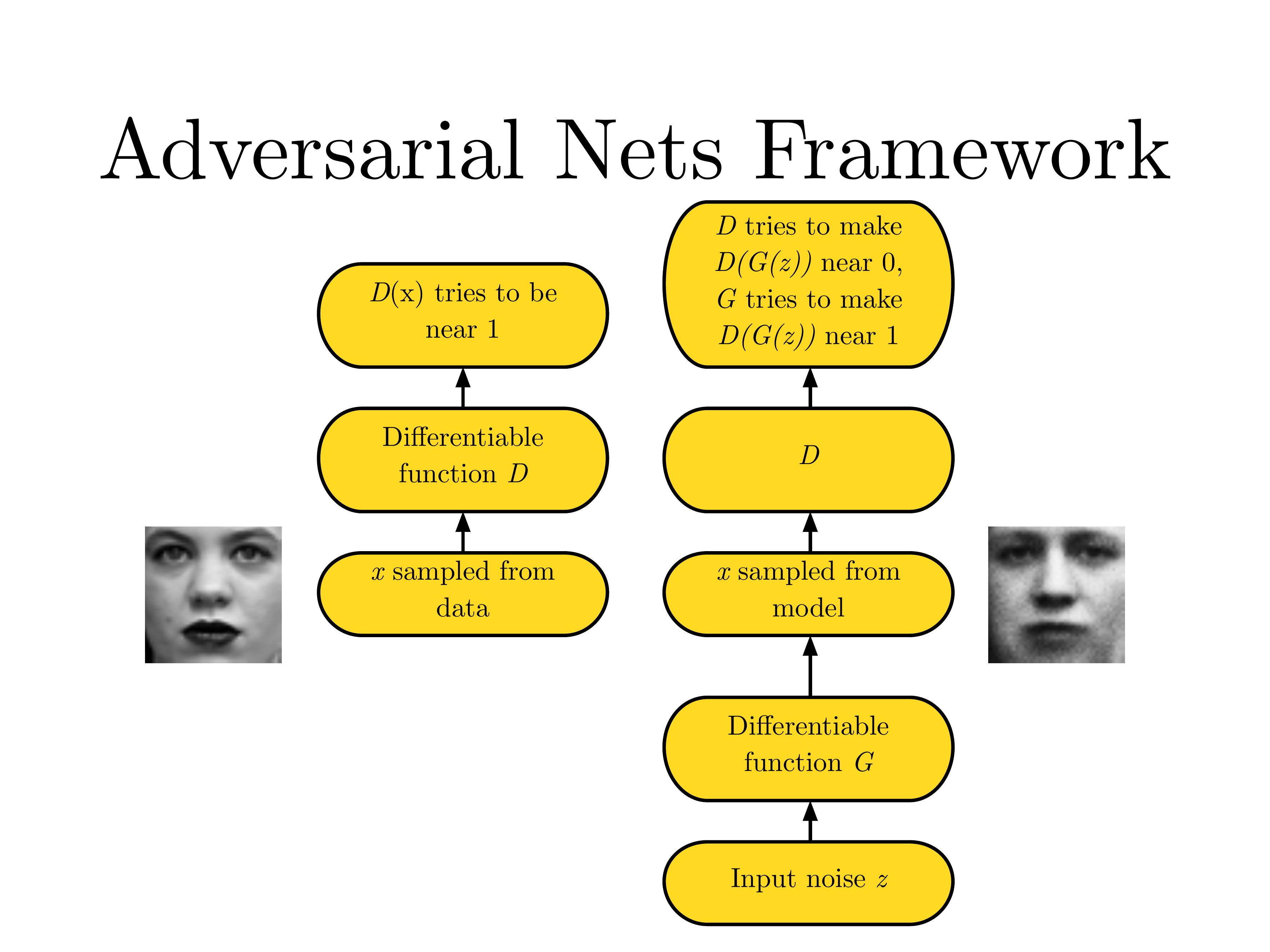}
\caption{The GAN framework pits two adversaries against each other in a game.
Each player is represented by a differentiable function controlled by a set
of parameters.
Typically these functions are implemented as deep neural networks.
The game plays out in two scenarios.
In one scenario, training examples $\vx$ are randomly sampled from the training
set and used as input for the first player, the discriminator, represented
by the function $D$. The goal of the discriminator is to output the probability
that its input is real rather than fake, under the assumption that half of the
inputs it is ever shown are real and half are fake.
In this first scenario, the goal of the discriminator is for $D(\vx)$ to be
near 1.
In the second scenario, inputs $\vz$ to the generator are randomly sampled from
the model's prior over the latent variables.
The discriminator then receives input $G(\vz)$, a fake sample created by the
generator.
In this scenario, both players participate. The discriminator strives to make
$D(G(z))$ approach 0 while the generative strives to make the same quantity
approach 1.
If both models have sufficient capacity, then the Nash equilibrium of this game
corresponds to the $G(\vz)$ being drawn from the same distribution as the training
data, and $D(\vx) = \frac{1}{2}$ for all $\vx$.
}
\label{fig:framework}
\end{figure}

The basic idea of GANs is to set up a game between two players.
One of them is called the \newterm{generator}.
The generator creates samples that are intended to come from the
same distribution as the training data.
The other player is the \newterm{discriminator}.
The discriminator examines samples to determine whether they are real
or fake.
The discriminator learns using traditional supervised learning techniques,
dividing inputs into two classes (real or fake).
The generator is trained to fool the discriminator.
We can think of the generator as being like a counterfeiter, trying to
make fake money, and the discriminator as being like police, trying to
allow legitimate money and catch counterfeit money.
To succeed in this game, the counterfeiter must learn to make money that
is indistinguishable from genuine money, and the generator network must
learn to create samples that are drawn from the same distribution as the
training data.
The process is illustrated in \figref{fig:framework}.

\begin{figure}
  \centering
  \includegraphics{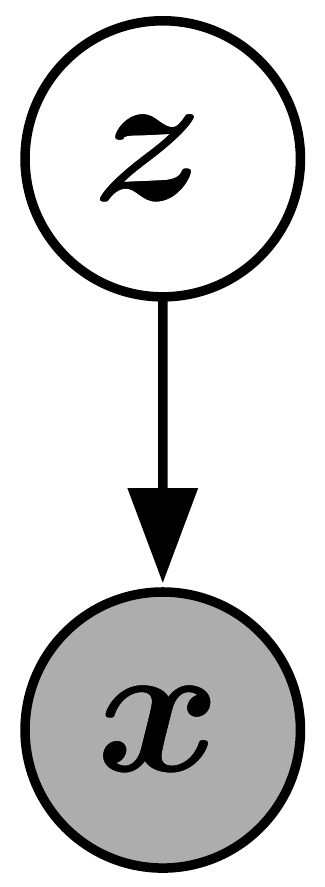}
  \caption{The graphical model structure of GANs, which is also shared with
    VAEs, sparse coding, etc.
    It is directed graphical model where every latent variable influences
    every observed variable.
    Some GAN variants remove some of these connections.
  }
  \label{fig:graph}
\end{figure}

Formally, GANs are a structured probabilistic model (see chapter 16 of
\citet{Goodfellow-et-al-2016} for an introduction to structured probabilistic
models) containing latent variables $\vz$ and observed variables $\vx$.
The graph structure is shown in \figref{fig:graph}.

The two players in the game are represented by two functions, each of which
is differentiable both with respect to its inputs and with respect to its
parameters.
The discriminator is a function $D$ that takes $\vx$ as input and uses
$\vtheta^{(D)}$ as parameters.
The generator is defined by a function $G$ that takes $\vz$ as input and
uses $\vtheta^{(G)}$ as parameters.

Both players have cost functions that are defined in terms of both players'
parameters.
The discriminator wishes to minimize $J^{(D)}\left( \vtheta^{(D)}, \vtheta^{(G)} \right)$
and must do so while controlling only $\vtheta^{(D)}$.
The generator wishes to minimize $J^{(G)}\left( \vtheta^{(D)}, \vtheta^{(G)} \right)$
and must do so while controlling only $\vtheta^{(G)}$.
Because each player's cost depends on the other player's parameters, but each player
cannot control the other player's parameters, this scenario is most
straightforward to describe as a game rather than as an optimization problem.
The solution to an optimization problem is a (local) minimum, a point in parameter space
where all neighboring points have greater or equal cost.
The solution to a game is a Nash equilibrium.
Here, we use the terminology of local differential Nash equilibria \citep{ratliff2013characterization}.
In this context, a Nash equilibrium is a tuple $(\vtheta^{(D)}, \vtheta^{(G)} )$
that is a local minimum of $J^{(D)}$ with respect to $\vtheta^{(D)}$ and a local
minimum of $J^{(G)}$ with respect to $\vtheta^{(G)}$.

\paragraph{The generator}
The generator is simply a differentiable function $G$.
When $\vz$ is sampled from some simple prior distribution,
$G(\vz)$ yields a sample of $\vx$ drawn from $\pmodel$.
Typically, a deep neural network is used to represent $G$.
Note that the inputs to the function $G$ do not need to correspond
to inputs to the first layer of the deep neural net; inputs may
be provided at any point throughout the network.
For example, we can partition $\vz$ into two vectors $\vz^{(1)}$
and $\vz^{(2)}$, then feed $\vz^{(1)}$ as input to the first layer
of the neural net and add $\vz^{(2)}$ to the last layer of the neural
net. If $\vz^{(2)}$ is Gaussian, this makes $\vx$ conditionally Gaussian
given $\vz^{(1)}$.
Another popular strategy is to apply additive or multiplicative noise to
hidden layers or concatenate noise to hidden layers of the neural net.
Overall, we see that there are very few restrictions on the design of
the generator net.
If we want $\pmodel$ to have full support on $\vx$ space we need the dimension
of $\vz$ to be at least as large as the dimension of $\vx$, and $G$ must be
differentiable, but those are the only requirements.
In particular, note that any model that can be trained with the nonlinear
ICA approach can be a GAN generator network.
The relationship with variational autoencoders is more complicated;
the GAN framework can train some models that the VAE framework cannot and vice
versa, but the two frameworks also have a large intersection.
The most salient difference is that, if relying on standard backprop,
VAEs cannot have discrete variables at the input to the generator,
while GANs cannot have discrete variables at the output of the generator.

\paragraph{The training process}
The training process consists of simultaneous SGD.
On each step, two minibatches are sampled: a minibatch of $\vx$ values from
the dataset and a minibatch of $\vz$ values drawn from the model's prior over
latent variables.
Then two gradient steps are made simultaneously:
one updating $\vtheta^{(D)}$ to reduce $J^{(D)}$
and one updating $\vtheta^{(G)}$ to reduce $J^{(G)}$.
In both cases, it is possible to use the gradient-based optimization algorithm
of your choice.
Adam \citep{kingma2014adam} is usually a good choice.
Many authors recommend running more steps of one player than the other, but
as of late 2016, the author's opinion is that the protocol that works the best
in practice is simultaneous gradient descent, with one step for each player.

\subsection{Cost functions}

Several different cost functions may be used within the GANs framework.

\subsubsection{The discriminator's cost, $J^{(D)}$}

All of the different games designed for GANs so far use the same cost for the
discriminator, $J^{(D)}$. They differ only in terms of the cost used for the
generator, $J^{(G)}$.

The cost used for the discriminator is:
\begin{equation}
  J^{(D)}(\vtheta^{(D)}, \vtheta^{(G)}) = -\frac{1}{2} \E_{\vx \sim \pdata} \log D(\vx) - \frac{1}{2} \E_{\vz} \log \left(1 - D\left( G(z)\right) \right).
  \label{eq:discriminator_cost}
\end{equation}

This is just the standard cross-entropy cost that is minimized when training a standard binary classifier
with a sigmoid output.
The only difference is that the classifier is trained on two minibatches of data; one coming from the
dataset, where the label is $1$ for all examples, and one coming from the generator, where the label
is $0$ for all examples.

All versions of the GAN game encourage the discriminator to minimize \eqref{eq:discriminator_cost}.
In all cases, the discriminator has the same optimal strategy.
The reader is now encouraged to complete the exercise in \secref{sec:opt_d} and review its solution
given in \secref{sec:opt_d_soln}. This exercise shows how to derive the optimal discriminator strategy
and discusses the importance of the form of this solution.

We see that by training the discriminator, we are able to obtain an estimate of the ratio
\[
  \frac{\pdata(\vx)}{\pmodel(\vx)}
\]
at every point $\vx$.
Estimating this ratio enables us to compute a wide variety of divergences and their gradients.
This is the key approximation technique that sets GANs apart from variational autoencoders
and Boltzmann machines.
Other deep generative models make approximations based on lower bounds or Markov chains;
GANs make approximations based on using supervised learning to estimate a ratio of two densities.
The GAN approximation is subject to the failures of supervised learning: overfitting and underfitting.
In principle, with perfect optimization and enough training data, these failures can be overcome.
Other models make other approximations that have other failures.

Because the GAN framework can naturally be analyzed with the tools of game theory,
we call GANs ``adversarial.'' But we can also think of them as cooperative, in
the sense that the discriminator estimates this ratio of densities and then freely
shares this information with the generator.
From this point of view, the discriminator is more like a teacher instructing the
generator in how to improve than an adversary.
So far, this cooperative view has not led to any particular change in the development
of the mathematics.

\subsubsection{Minimax}
\label{sec:minimax}

So far we have specified the cost function for only the discriminator.
A complete specification of the game requires that we specify a cost function also
for the generator.

The simplest version of the game is a \newterm{zero-sum game}, in which the sum of all player's
costs is always zero.
In this version of the game,
\begin{equation}
J^{(G)} = - J^{(D)}.
\label{eq:minimax}
\end{equation}

Because $J^{(G)}$ is tied directly to $J^{(D)}$, we can summarize the entire game with a
\newterm{value function} specifying the discriminator's payoff:
\[ V\left(\vtheta^{(D)}, \vtheta^{(G)} \right) = - J^{(D)} \left(\vtheta^{(D)}, \vtheta^{(G)} \right).\]

Zero-sum games are also called \newterm{minimax} games because their solution involves minimization
in an outer loop and maximization in an inner loop:
\[ \vtheta^{(G)*} = \argmin_{\vtheta^{(G)}} \max_{\vtheta^{(D)}} V\left(\vtheta^{(D)}, \vtheta^{(G)} \right) . \]

The minimax game is mostly of interest because it is easily amenable to theoretical analysis.
\citet{Goodfellow-et-al-NIPS2014-small} used this variant of the GAN game to show that learning in
this game resembles minimizing the Jensen-Shannon divergence between the data and the model distribution,
and that the game converges to its equilibrium if both players' policies can be updated directly in
function space.
In practice, the players are represented with deep neural nets and updates are made in parameter space,
so these results, which depend on convexity, do not apply.

\subsubsection{Heuristic, non-saturating game}
\label{sec:heuristic}

The cost used for the generator in the minimax game (\eqref{eq:minimax}) is useful for theoretical analysis,
but does not perform especially well in practice.

Minimizing the cross-entropy between a target class and a classifier's predicted distribution
is highly effective because the cost never saturates when the classifier has the wrong output.
The cost does eventually saturate, approaching zero, but only when the classifier has already
chosen the correct class.

In the minimax game, the discriminator minimizes a cross-entropy, but the generator maximizes
the same cross-entropy.
This is unfortunate for the generator, because when the discriminator successfully rejects
generator samples with high confidence, the generator's gradient vanishes.

To solve this problem, one approach is to continue to use cross-entropy minimization for the
generator.
Instead of flipping the sign on the discriminator's cost to obtain a cost for the generator,
we flip the target used to construct the cross-entropy cost.
The cost for the generator then becomes:
\[
  J^{(G)} = -\frac{1}{2} \E_\vz \log D(G(\vz))
\]

In the minimax game, the generator minimizes the log-probability of the discriminator being correct.
In this game, the generator maximizes the log-probability of the discriminator being mistaken.

This version of the game is heuristically motivated, rather than being motivated by a theoretical
concern.
The sole motivation for this version of the game is to ensure that each player has a strong
gradient when that player is ``losing'' the game.

In this version of the game, the game is no longer zero-sum, and cannot be described with a single
value function.

\subsubsection{Maximum likelihood game}
\label{sec:mle_gan}

We might like to be able to do maximum likelihood learning with GANs, which would mean minimizing
the KL divergence between the data and the model, as in \eqref{eq:kl}.
Indeed, in \secref{sec:tree}, we said that GANs could optionally implement maximum likelihood,
for the purpose of simplifying their comparison to other models.

There are a variety of methods of approximating \eqref{eq:kl} within the GAN
framework.
\citet{Goodfellow-ICLR2015} showed that using
\[ J^{(G)} = -\frac{1}{2} \E_z \exp\left( \sigma^{-1} \left( D(G(\vz)) \right) \right), \]
  where $\sigma$ is the logistic sigmoid function, is equivalent to minimizing \eqref{eq:kl},
  under the assumption that the discriminator is optimal.
  This equivalence holds in expectation; in practice, both stochastic gradient descent on the KL
  divergence and the GAN training procedure will have some variance around the true expected
  gradient due to the use of sampling (of $\vx$ for maximum likelihood and $\vz$ for GANs)
  to construct the estimated gradient.
  The demonstration of this equivalence is an exercise (\secref{sec:mle_exercise}
  with the solution in \secref{sec:mle_soln}).

  Other methods of approximating maximum likelihood within the GANs framework are possible.
  See for example \citet{nowozin2016f}.

  \subsubsection{Is the choice of divergence a distinguishing feature of GANs?}
  \label{sec:which_divergence}

  As part of our investigation of how GANs work, we might wonder exactly what it is
  that makes them work well for generating samples.

  Previously, many people (including the author) believed that GANs produced sharp,
  realistic samples because they minimize the Jensen-Shannon divergence while
  VAEs produce blurry samples because they minimize the KL divergence between the
  data and the model.

\begin{figure}
    \centering
    \includegraphics[width=\figwidth]{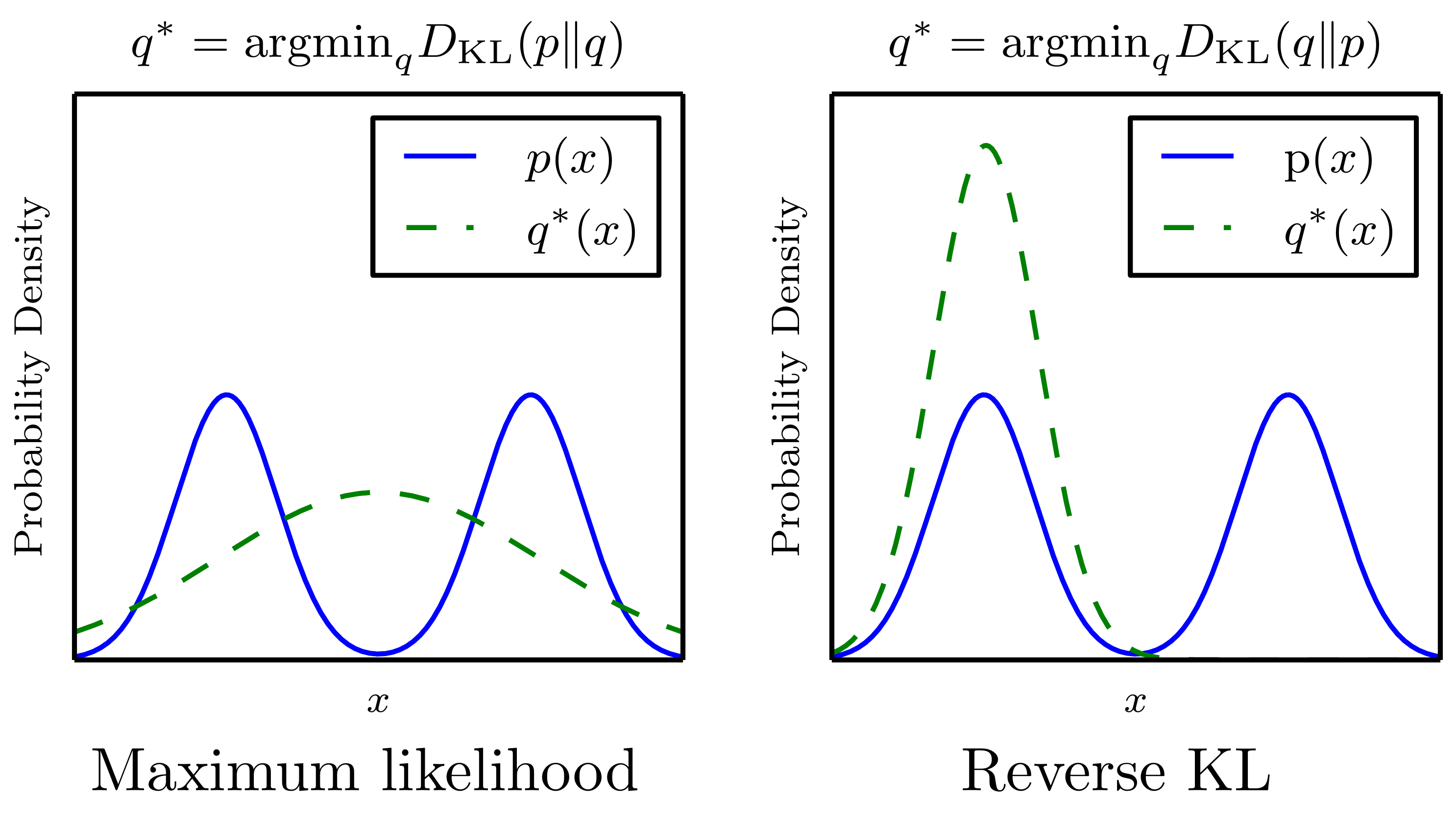}
    \caption{
The two directions of the KL divergence are not equivalent.
The differences are most obvious when the model has too little
capacity to fit the data distribution.
Here we show an example of a distribution over one-dimensional
data $x$.
In this example, we use a mixture of two Gaussians as the data
distribution, and a single Gaussian as the model family.
Because a single Gaussian cannot capture the true data distribution,
the choice of divergence determines the tradeoff that the model makes.
On the left, we use the maximum likelihood criterion.
The model chooses to average out the two modes, so that it places high
probability on both of them.
On the right, we use the reverse order of the arguments to the KL
divergence, and the model chooses to capture only one of the two modes.
It could also have chosen the other mode; the two are both local minima
of the reverse KL divergence.
We can think of $\KL( \pdata \Vert \pmodel )$ as preferring to place
high probability everywhere that the data occurs,
and $\KL( \pmodel \Vert \pdata )$ as preferrring to place low probability
wherever the data does not occur.
From this point of view, one might expect
$\KL( \pmodel \Vert \pdata )$
to yield more visually pleasing samples,
because the model will not choose to generate unusual samples lying
between modes of the data generating distribution.
    }
    \label{fig:kl}
  \end{figure}

  The KL divergence is not symmetric; minimizing $\KL(\pdata \Vert \pmodel)$
  is different from minimizing $\KL(\pmodel \Vert \pdata)$.
  Maximum likelihood estimation performs the former; minimizing the Jensen-Shannon
  divergence is somewhat more similar to the latter.
  As shown in \figref{fig:kl}, the latter might be expected to yield better samples
  because a model trained with this divergence would prefer
  to generate samples that come only from modes in the training distribution
  even if that means ignoring some modes, rather than
  including all modes but generating some samples that do not come 
  from any training set mode.

\begin{figure}
\centering
\includegraphics[width=\figwidth]{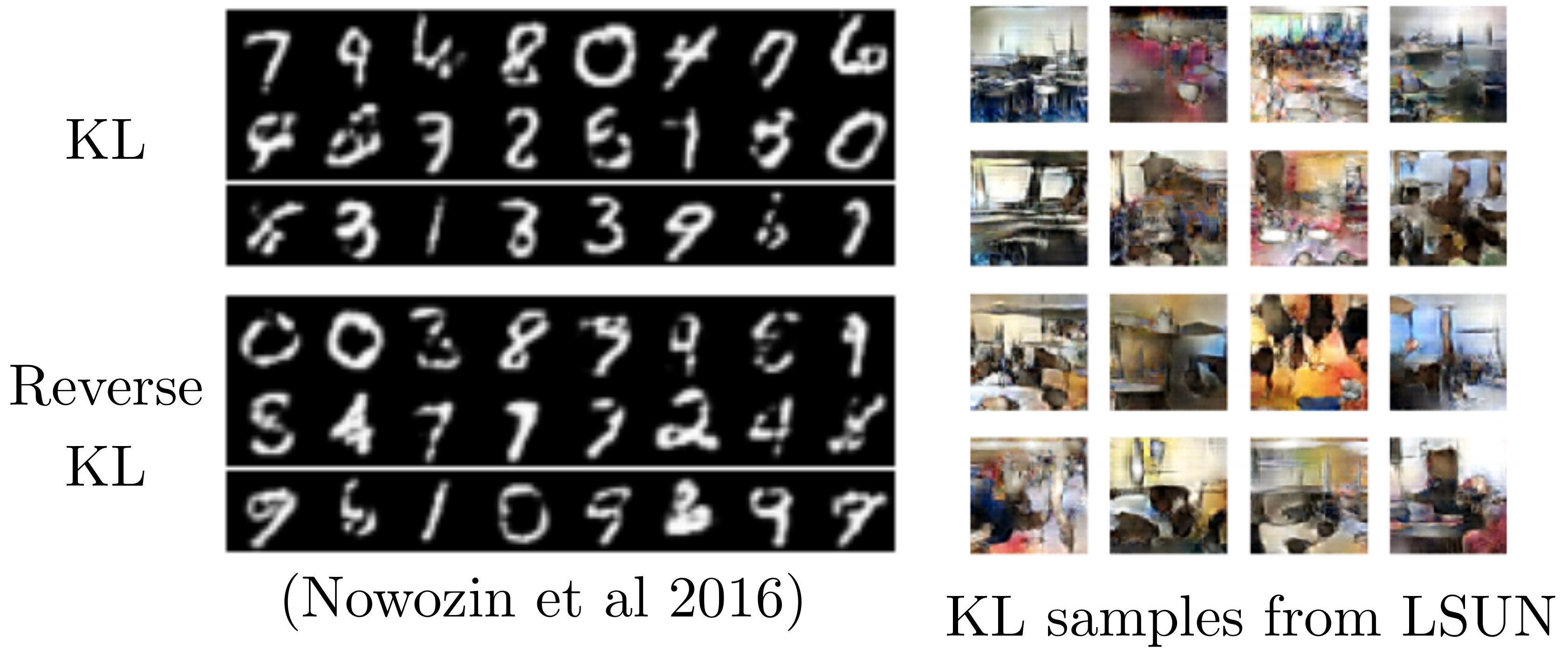}
\caption{
The f-GAN model is able to minimize many different divergences.
Because models trained to minimize $\KL(\pdata \Vert \pmodel)$
still generate sharp samples and tend to select a small number
of modes, we can conclude that the use of the Jensen-Shannon
divergence is not a particular important distinguishing
characteristic of GANs and that it does not explain why their
samples tend to be sharp.
}
\label{fig:fgan}
\end{figure}

Some newer evidence suggests that the use of the Jensen-Shannon divergence does
not explain why GANs make sharper samples:
\begin{itemize}
  \item It is now possible to train GANs using maximum likelihood, as described in
    \secref{sec:mle_gan}.
    These models still generate sharp samples, and still select a small number of modes.
    See \figref{fig:fgan}.
  \item
    GANs often choose to generate from very few modes; fewer than the limitation
    imposed by the model capacity.
    The reverse KL prefers to generate from {\em as many modes of the data distribution as the model is able to};
    it does not prefer fewer modes in general.
    This suggests that the mode collapse is driven by a factor other than the choice 
    of divergence.
\end{itemize}

Altogether, this suggests that GANs choose to generate a small number of modes
due to a defect in the training procedure, rather than due to the divergence
they aim to minimize.
This is discussed further in \secref{sec:mode_collapse}.
The reason that GANs produce sharp samples is not entirely clear.
It may be that the family of models trained using GANs is different from the family
of models trained using VAEs (for example, with GANs it is easy to make models
where $\vx$ has a more complicated distribution than just an isotropic Gaussian
conditioned on the input to the generator).
It may also be that the kind of approximations that GANs make have different
effects than the kind of approximations that other frameworks make.

\subsubsection{Comparison of cost functions}

We can think of the generator network as learning by a strange kind of reinforcement
learning.
Rather than being told a specific output $\vx$ it should associate with each $\vz$,
the generator takes actions and receives rewards for them.
In particular, note that $J^{(G)}$ does not make reference to the training data
directly at all; all information about the training data comes only through what
the discriminator has learned. (Incidentally, this makes GANs resistant to overfitting,
because the generator has no opportunity in practice to directly copy training examples)
The learning process differs somewhat from traditional reinforcement learning because
\begin{itemize}
 \item The generator is able to observe not just the output of the reward function but
   also its gradients.
 \item The reward function is non-stationary; the reward is based on the discriminator
   which learns in response to changes in the generator's policy.
   \end{itemize}

In all cases, we can think of the sampling process that begins with the selection of
a specific $\vz$ value as an episode that receives a single reward, independent of the
actions taken for all other $\vz$ values.
The reward given to the generator is a function of a single scalar value,
$D(G(\vz))$.
We usually think of this in terms of cost (negative reward).
The cost for the generator is always monotonically decreasing in $D(G(\vz))$ but
different games are designed to make this cost decrease faster along different parts of the curve.

\begin{figure}
\centering
\includegraphics[width=\figwidth]{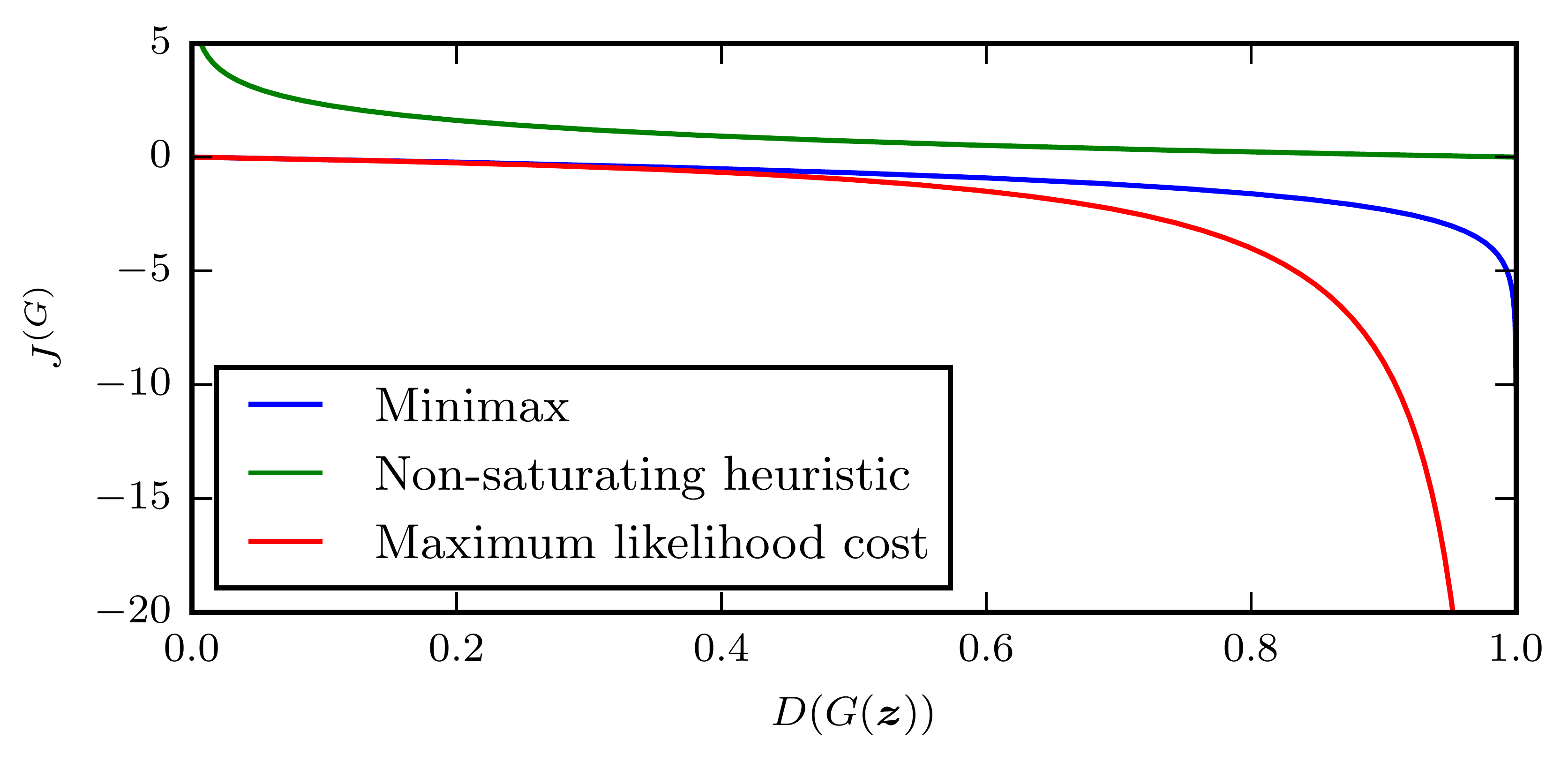}
\caption{
  The cost that the generator receives for generating a samples $G(\vz)$ depends only on
  how the discriminator responds to that sample.
  The more probability the discriminator assigns to the sample being real, the less cost
  the generator receives.
  We see that when the sample is likely to be fake, both the minimax game and the maximum
  likelihood game have very little gradient, on the flat left end of the curve.
  The heuristically motivated non-saturating cost avoids this problem.
  Maximum likelihood also suffers from the problem that nearly all of the gradient comes
  from the right end of the curve, meaning that a very small number of samples dominate
  the gradient computation for each minibatch.
  This suggests that variance reduction techniques could be an important research area
  for improving the performance of GANs, especially GANs based on maximum likelihood.
  Figure reproduced from \citet{Goodfellow-ICLR2015}.
}
\label{fig:comparison}
\end{figure}

\Figref{fig:comparison} shows the cost response curves as functions of $D(G(\vz))$ for three
different variants of the GAN game.
We see that the maximum likelihood game gives very high variance in the cost, with most of the
cost gradient coming from the very few samples of $\vz$ that correspond to the samples that
are most likely to be real rather than fake.
The heuristically designed non-saturating cost has lower sample variance, which may explain
why it is more successful in practice.
  This suggests that variance reduction techniques could be an important research area
  for improving the performance of GANs, especially GANs based on maximum likelihood.

\subsection{The DCGAN architecture}

\begin{figure}
\centering
\includegraphics[width=\textwidth]{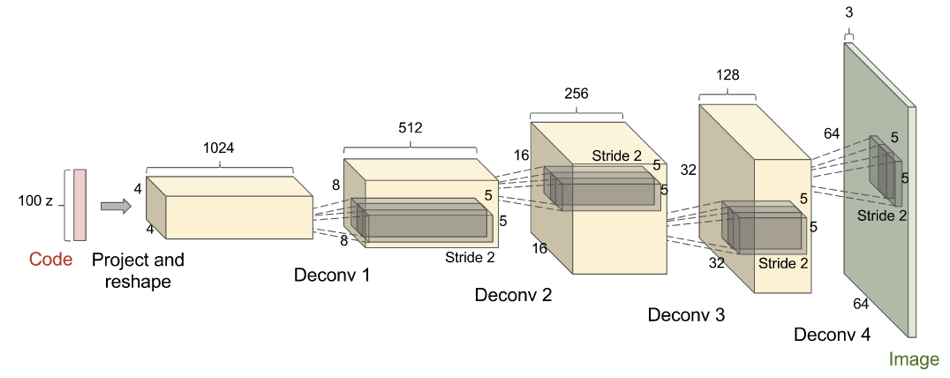}
  \caption{The generator network used by a DCGAN. Figure reproduced from \citet{radford2015unsupervised}.}
\label{fig:dcgan}
\end{figure}

Most GANs today are at least loosely based on the DCGAN architecture \citep{radford2015unsupervised}.
DCGAN stands for ``deep, convolution GAN.'' Though GANs were both deep and convolutional prior to
DCGANs, the name DCGAN is useful to refer to this specific style of architecture.
Some of the key insights of the DCGAN architecture were to:
\begin{itemize}
  \item Use batch normalization \citep{Ioffe+Szegedy-2015} layers in most layers of both the discriminator and the generator,
        with the two minibatches for the discriminator normalized separately.
        The last layer of the generator and first layer of the discriminator are not batch normalized,
        so that the model can learn the correct mean and scale of the data distribution.
        See \figref{fig:dcgan}.
  \item The overall network structure is mostly borrowed from the all-convolutional net \citep{Springenberg2015}.
        This architecture contains neither pooling nor ``unpooling'' layers.
        When the generator needs to increase the spatial dimension of the representation
        it uses transposed convolution with a stride greater than 1.
  \item The use of the Adam optimizer rather than SGD with momentum.
\end{itemize}

\begin{figure}
  \centering
  \includegraphics[width=\figwidth]{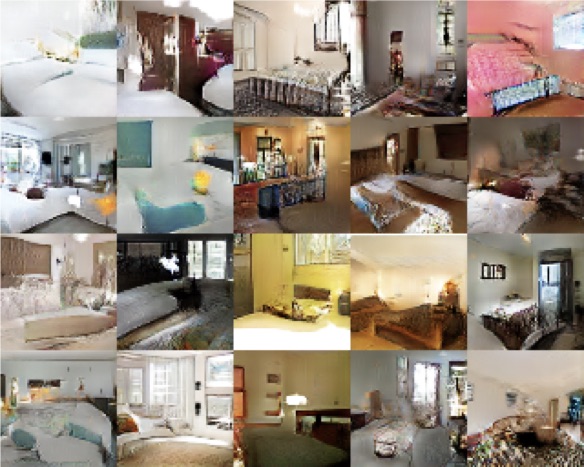}
  \caption{Samples of images of bedrooms generated by a DCGAN trained on the LSUN dataset.}
  \label{fig:dcgan_lsun}
\end{figure}

\begin{figure}
\centering
$
\vcenter{
    \hbox{%
\includegraphics[width=.15\figwidth]{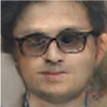} %
    }
}
\vcenter{\hbox{-}} %
\vcenter{\hbox{
\includegraphics[width=.15\figwidth]{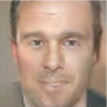} %
}}
\vcenter{\hbox{+}} %
\vcenter{\hbox{
\includegraphics[width=.15\figwidth]{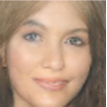} %
}}
\vcenter{\hbox{=}} %
\vcenter{\hbox{
\includegraphics[width=.45\figwidth]{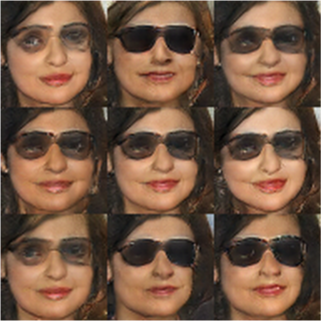}
}
}
$
\caption{DCGANs demonstrated that GANs can learn a distributed representation
that disentangles the concept of gender from the concept of wearing
glasses. If we begin with the representation of the concept of a
man with glasses, then subtract the vector representing the concept
of a man without glasses, and finally add the vector representing
the concept of a woman without glasses, we obtain the vector representing
the concept of a woman with glasses. The generative model correctly
decodes all of these representation vectors to images that may be
recognized as belonging to the correct class.
Images reproduced from \citet{radford2015unsupervised}.
}
\label{fig:dcgan_face_arithmetic}
\end{figure}

Prior to DCGANs, LAPGANs \citep{denton2015deep} were the only version of GAN
that had been able to scale to high resolution images.
LAPGANs require a multi-stage generation process in which multiple GANs
generate different levels of detail in a Laplacian pyramid representation
of an image.
DCGANs were the first GAN model to learn to generate high resolution images
in a single shot.
As shown in \figref{fig:dcgan_lsun}, DCGANs are able to generate high quality
images when trained on restricted domains of images, such as images of bedrooms.
DCGANs also clearly demonstrated that GANs learn to use their latent code
in meaningful ways, with simple arithmetic operations in latent space
having clear interpretation as arithmetic operations on semantic attributes
of the input, as demonstrated in \figref{fig:dcgan_face_arithmetic}.

\subsection{How do GANs relate to noise-contrastive estimation and maximum likelihood?}

\begin{figure}
\centering
\includegraphics[width=\figwidth]{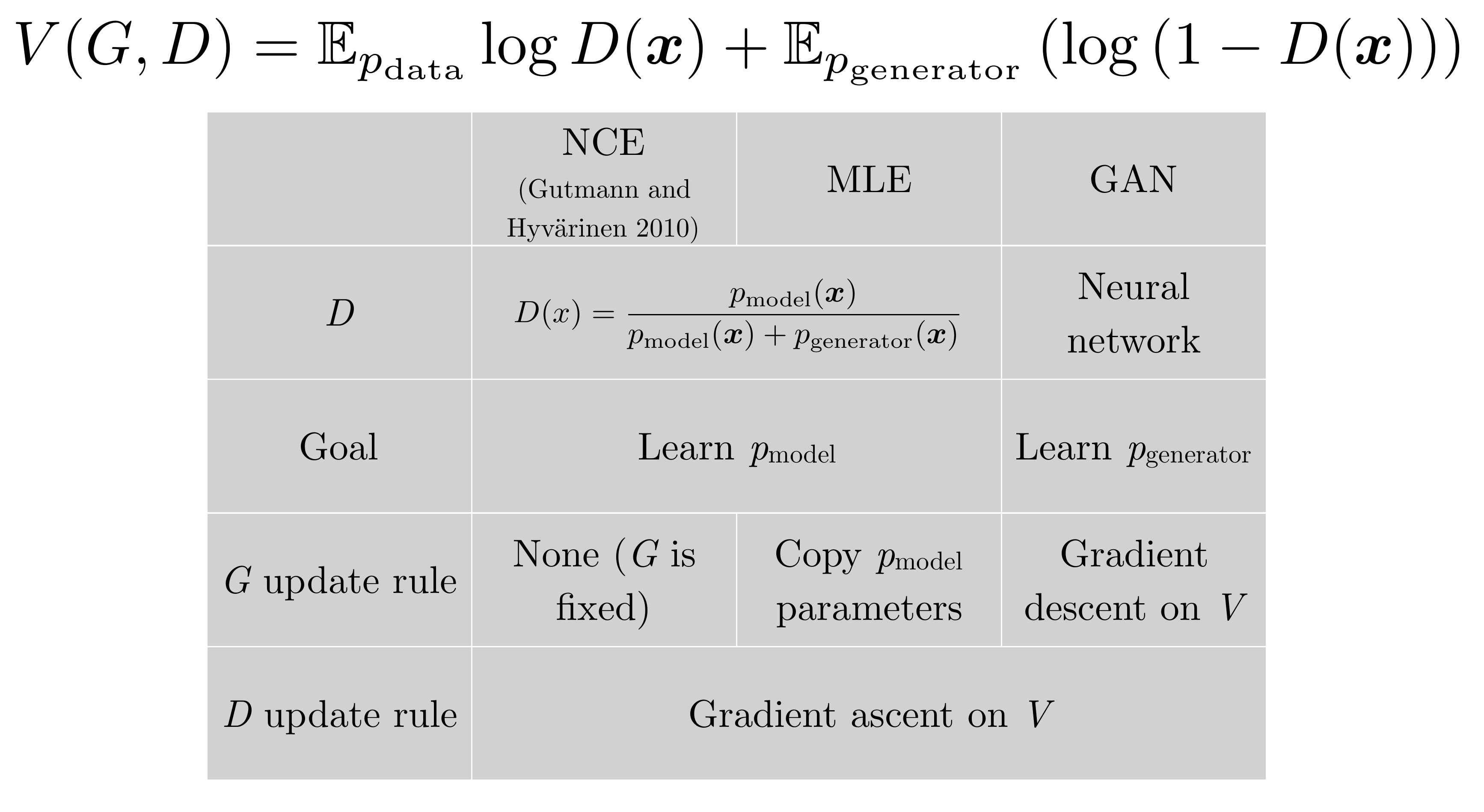}
\caption{
  \citet{Goodfellow-ICLR2015} demonstrated the following
  connections between minimax GANs, noise-contrastive estimation, and
  maximum likelihood: all three can be interpreted as strategies
  for playing a minimax game with the same value function.
  The biggest difference is in where $\pmodel$ lies.
  For GANs, the generator is $\pmodel$, while for NCE and MLE,
  $\pmodel$ is part of the discriminator.
  Beyond this, the differences between the methods lie in the update
  strategy.
  GANs learn both players with gradient descent.
  MLE learns the discriminator using gradient descent, but has a heuristic
  update rule for the generator.
  Specifically, after each discriminator update step, MLE copies the density model learned inside the discriminator and 
  converts it into a sampler to be used as the generator.
  NCE never updates the generator; it is just a fixed source of noise.
}
\label{fig:nce}
\end{figure}

While trying to understand how GANs work, one might naturally wonder about
how they are connected to \newterm{noise-constrastive estimation} (NCE)
\citep{Gutmann+Hyvarinen-2010}.
Minimax GANs use the cost function from NCE as their value function, so the
methods seem closely related at face value.
It turns out that they actually learn very different things, because the
two methods focus on different players within this game.
Roughly speaking, the goal of NCE is to learn the density model within the
discriminator, while the goal of GANs is to learn the sampler defining
the generator.
While these two tasks seem closely related at a qualitative level, the
gradients for the tasks are actually quite different.
Surprisingly, maximum likelihood turns out to be closely related to NCE,
and corresponds to playing a minimax game with the same value function,
but using a sort of heuristic update strategy rather than gradient descent
for one of the two players.
The connections are summarized in \figref{fig:nce}.

\section{Tips and Tricks}

Practitioners use several tricks to improve the performance of GANs.
It can be difficult to tell how effective some of these tricks are;
many of them seem to help in some contexts and hurt in others.
These should be regarded as techniques that are worth trying out,
not as ironclad best practices.

NIPS 2016 also featured a workshop on adversarial training, with
an invited talk by Soumith Chintala called "How to train a GAN."
This talk has more or less the same goal as this portion of the tutorial,
with a different collection of advice.
To learn about tips and tricks not included in this tutorial, check
out the GitHub repository associated with Soumith's talk:

\url{https://github.com/soumith/ganhacks}

\subsection{Train with labels}

Using labels in any way, shape or form almost always results in a dramatic
improvement in the subjective quality of the samples generated by the model.
This was first observed by \citet{denton2015deep}, who built class-conditional
GANs that generated much better samples than GANs that were free to generate
from any class.
Later, \citet{salimans2016improved} found that sample quality improved
even if the generator did not explicitly incorporate class information; training
 the discriminator to recognize specific classes of real objects is sufficient.

 It is not entirely clear why this trick works.
 It may be that the incorporation of class information gives the training
 process useful clues that help with optimization.
 It may also be that this trick gives no objective improvement in sample quality,
 but instead biases the samples toward taking on properties that the human
 visual system focuses on.
 If the latter is the case, then this trick may not result in a better model
 of the true data-generating distribution, but it still helps to create media
 for a human audience to enjoy and may help an RL agent to carry out tasks
 that rely on knowledge of the same aspects of the environment that are relevant
 to human beings.

 It is important to compare results obtained using this trick only to other
 results using the same trick; models trained with labels should be compared
 only to other models trained with labels, class-conditional models should
 be compared only to other class-conditional models.
 Comparing a model that uses labels to one that does not is unfair and an
 uninteresting benchmark, much as a convolutional model can usually be expected
 to outperform a non-convolutional model on image tasks.

\subsection{One-sided label smoothing}
\label{sec:label_smooth}

GANs are intended to work when the discriminator estimates a ratio of two
densities, but deep neural nets are prone to producing highly confident
outputs that identify the correct class but with too extreme of a probability.
This is especially the case when the input to the deep network is adversarially
constructed; the classifier tends to linearly extrapolate and produce
extremely confident predictions \citep{Goodfellow-2015-adversarial}.

To encourage the discriminator to estimate soft probabilities rather than
to extrapolate to extremely confident classification, we can use a technique
called \newterm{one-sided label smoothing} \citep{salimans2016improved}.

Usually we train the discriminator using \eqref{eq:discriminator_cost}.
We can write this in TensorFlow \citep{tensorflow} code as:
\begin{lstlisting}
d_on_data = discriminator_logits(data_minibatch)
d_on_samples = discriminator_logits(samples_minibatch)
loss = tf.nn.sigmoid_cross_entropy_with_logits(d_on_data, 1.) + \
       tf.nn.sigmoid_cross_entropy_with_logits(d_on_samples, 0.)
\end{lstlisting}

The idea of one-sided label smoothing is to replace the target for the real examples
with a value slightly less than one, such as .9:

\begin{lstlisting}
loss = tf.nn.sigmoid_cross_entropy_with_logits(d_on_data, .9) + \
       tf.nn.sigmoid_cross_entropy_with_logits(d_on_samples, 0.)
\end{lstlisting}

This prevents extreme extrapolation behavior in the discriminator; if it learns
to predict extremely large logits corresponding to a probability approaching $1$
for some input, it will be penalized and encouraged to bring the logits back
down to a smaller value.

It is important to not smooth the labels for the fake samples.
Suppose we use a target of $1-\alpha$ for the real data and a target of $0+\beta$
for the fake samples.
Then the optimal discriminator function is
\[ D^*(\vx) = \frac{(1-\alpha) \pdata(\vx) + \beta \pmodel(\vx)} { \pdata(\vx) + \pmodel(\vx) }.\]

When $\beta$ is zero, then smoothing by $\alpha$ does nothing but scale down the optimal value
of the discriminator.
When $\beta$ is nonzero, the shape of the optimal discriminator function changes.
In particular, in a region where $\pdata(\vx)$ is very small and $\pmodel(\vx)$ is larger,
$D^*(\vx)$ will have a peak near the spurious mode of $\pmodel(\vx)$.
The discriminator will thus reinforce incorrect behavior in the generator; the generator
will be trained either to produce samples that resemble the data or to produce samples
that resemble the samples it already makes.

One-sided label smoothing is a simple modification of the much older label smoothing
technique, which dates back to at least the 1980s.
\citet{Szegedy-et-al-2015} demonstrated that label smoothing is an excellent regularizer
in the context of convolutional networks for object recognition.
One reason that label smoothing works so well as a regularizer is that it does not
ever encourage the model to choose an incorrect class on the training set, but only
to reduce the confidence in the correct class.
Other regularizers such as weight decay often encourage some misclassification
if the coefficient on the regularizer is set high enough.
\citet{wardefarley2016} showed that label smoothing can help to reduce vulnerability to
adversarial examples, which suggests that label smoothing should help the discriminator
more efficiently learn to resist attack by the generator.

\subsection{Virtual batch normalization}

Since the introduction of DCGANs, most GAN architectures have involved some form
of batch normalization.
The main purpose of batch normalization is to improve the optimization of the model,
by reparameterizing the model so that the mean and variance of each feature are determined
by a single mean parameter and a single variance parameter associated with that feature,
rather than by a complicated interaction between all of the weights of all of the layers
used to extract the feature.
This reparameterization is accomplished by subtracting the mean and dividing by the standard
deviation of that feature on a minibatch of data.
It is important that the normalization operation is {\em part of the model},
so that back-propgation computes the gradient of features that are defined to always
be normalized.
The method is much less effect if features are frequently renormalized after learning
without the normalization defined as part of the model.

\begin{figure}
\centering
\includegraphics[width=\figwidth]{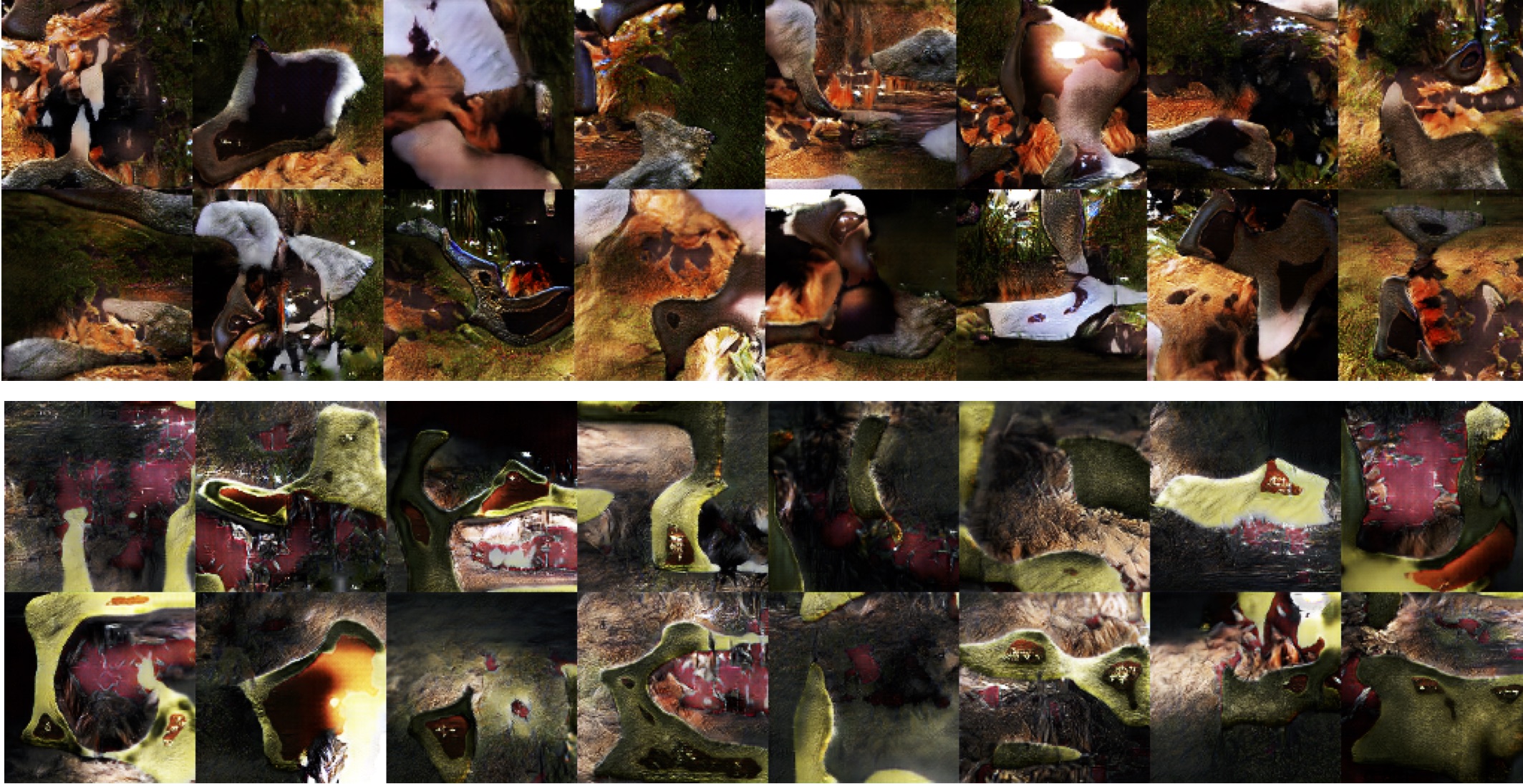}
\caption{
Two minibatches of sixteen samples each, generated by a generator network using
batch normalization.
These minibatches illustrate a problem that occurs occasionally when using batch
normalization: fluctuations in the mean and standard deviation of feature values
in a minibatch can have a greater effect than the individual $\vz$ codes for individual
images within the minibatch.
This manifests here as one minibatch containing all orange-tinted samples and the other
containing all green-tinted samples.
The examples within a minibatch should be independent from each other, but in this
case, batch normalization has caused them to become correlated with each other.
}
\label{fig:bad_batchnorm}
\end{figure}

Batch normalization is very helpful, but for GANs has a few unfortunate side effects.
The use of a different minibatch of data to compute the normalization statistics
on each step of training results in fluctuation of these normalizing constants.
When minibatch sizes are small (as is often the case when trying to fit a large generative
model into limited GPU memory) these fluctuations can become large enough that they
have more effect on the image generated by the GAN than the input $\vz$ has.
See \figref{fig:bad_batchnorm} for an example.

\citet{salimans2016improved} introduced techniques to mitigate this problem.
\newterm{Reference batch normalization} consists of running the network twice:
once on a minibatch of \newterm{reference examples} that are sampled once at the
start of training and never replaced, and once on the current minibatch of examples
to train on.
The mean and standard deviation of each feature are computed using the reference
batch. The features for both batches are then normalized using these computed statistics.
A drawback to reference batch normalization is that the model can overfit to the
reference batch. To mitigate this problem slightly, one can instead use
\newterm{virutal batch normalization}, in which the normalization statistics for each
example are computed using the union of that example and the reference batch.
Both reference batch normalization and virtual batch normalization have the property
that all examples in the training minibatch are processed independently from each other,
and all samples produced by the generator (except those defining the reference batch)
are i.i.d.

\subsection{Can one balance $G$ and $D$?}

Many people have an intuition that it is necessary to somehow balance the two players
to prevent one from overpowering the other.
If such balance is desirable and feasible, it has not yet been demonstrated in any
compelling fashion.

The author's present belief is that GANs work by estimating the ratio of the data density
and model density. This ratio is estimated correctly only when the discriminator is
optimal, so it is fine for the discriminator to overpower the generator.

Sometimes the gradient for the generator can vanish when the discriminator becomes
too accurate.
The right way to solve this problem is not to limit the power of the discriminator,
but to use a parameterization of the game where the gradient does not vanish
(\secref{sec:heuristic}).

Sometimes the gradient for the generator can become very large if the discriminator
becomes too confident. Rather than making the discriminator less accurate, a better
way to resolve this problem is to use one-sided label smoothing (\secref{sec:label_smooth}).

The idea that the discriminator should always be optimal in order to best estimate
the ratio would suggest training the discriminator for $k > 1$ steps every time
the generator is trained for one step. In practice, this does not usually result in a
clear improvement.

One can also try to balance the generator and discriminator by choosing the model
size.
In practice, the discriminator is usually deeper and sometimes has more filters
per layer than the generator.
This may be because it is important for the discriminator to be able to correctly
estimate the ratio between the data density and generator density, but it may
also be an artifact of the mode collapse problem---since the generator tends not
to use its full capacity with current training methods, practitioners presumably
do not see much of a benefit from increasing the generator capacity.
If the mode collapse problem can be overcome, generator sizes will presumably
increase. It is not clear whether discriminator sizes will increase proportionally.

\section{Research Frontiers}

GANs are a relatively new method, with many research directions still
remaining open.

\subsection{Non-convergence}

The largest problem facing GANs that researchers should try to resolve is the issue
of non-convergence.

Most deep models are trained using an optimization algorithm that seeks out a low
value of a cost function.
While many problems can interfere with optimization, optimization algorithms usually
make reliable downhill progress.
GANs require finding the equilibrium to a game with two players.
Even if each player successfully moves downhill on that player's update,
the same update might move the other player uphill.
Sometimes the two players eventually reach an equilibrium, but in other scenarios
they repeatedly undo each others' progress without arriving anywhere useful.
This is a general problem with games not unique to GANs, so a general solution
to this problem would have wide-reaching applications.

To gain some intuition for how gradient descent performs when applied to games
rather than optimization, the reader is encouraged to solve the exercise in
\secref{sec:xy_exercise} and review its solution in \secref{sec:xy_soln} now.

Simultaneous gradient descent converges for some games but not all of them.

In the case of the minimax GAN game (\secref{sec:minimax}),
\citet{Goodfellow-et-al-NIPS2014-small} showed that simultaneous gradient
descent converges {\em if the updates are made in function space}.
In practice, the updates are made in parameter space, so the convexity
properties that the proof relies on do not apply.
Currently, there is neither a theoretical argument that GAN games should
converge when the updates are made to parameters of deep neural networks,
nor a theoretical argument that the games should not converge.

In practice, GANs often seem to oscillate, somewhat like what happens in
the toy example in \secref{sec:xy_soln}, meaning that they progress from
generating one kind of sample to generating another kind of sample without
eventually reaching an equilibrium.

Probably the most common form of harmful non-convergence encountered in the GAN
game is mode collapse.

\subsubsection{Mode collapse}
\label{sec:mode_collapse}

\begin{figure}
\centering
\includegraphics[width=\figwidth]{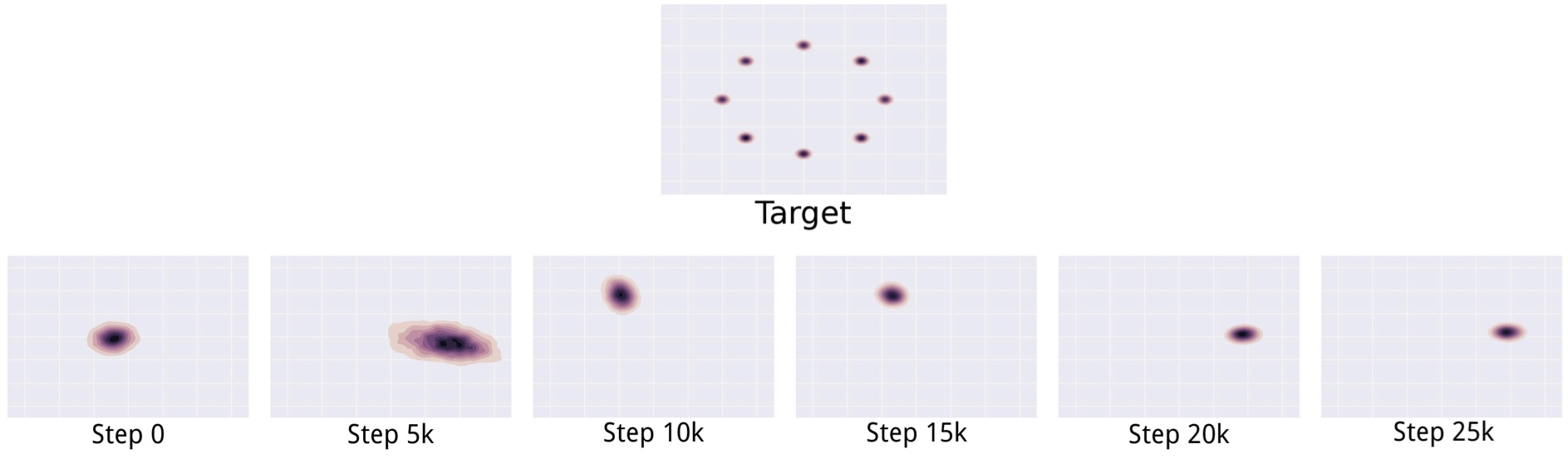}
\caption{
An illustration of the mode collapse problem on a two-dimensional toy dataset.
In the top row, we see the target distribution $\pdata$ that the model should
learn. It is a mixture of Gaussians in a two-dimensional space.
In the lower row, we see a series of different distributions learned over time
as the GAN is trained.
Rather than converging to a distribution containing all of the modes in the
training set, the generator only ever produces a single mode at a time, cycling
between different modes as the discriminator learns to reject each one.
Images from \citet{metz2016unrolled}.
}
\label{fig:mode_collapse}
\end{figure}

Mode collapse, also known as \newterm{the Helvetica scenario}, is a problem that occurs
when the generator learns to map several different input $\vz$ values to the same output
point.
In practice, complete mode collapse is rare, but partial mode collapse is common.
Partial mode collapse refers to scenarios in which the
generator makes multiple images that contain the same color or texture themes,
or multiple images containing different views of the same dog.
The mode collapse problem is illustrated in \figref{fig:mode_collapse}.

Mode collapse may arise because the maximin solution to the GAN game is different
from the minimax solution.
When we find the model
\[ G^* = \min_G \max_D V(G,D), \]
$G^*$ draws samples from the data distribution.
When we exchange the order of the min and max and find
\[ G^* = \max_D \min_G V(G, D), \]
the minimization with respect to the generator now lies in the inner
loop of the optimization procedure.
The generator is thus asked to map every $\vz$ value to the single $\vx$
coordinate that the discriminator believes is most likely to be real rather than fake.
Simultaneous gradient descent does not clearly privilege $\min \max$ over $\max \min$
or vice versa. We use it in the hope that it will behave like $\min \max$ but it
often behaves like $\max \min$.

As discussed in \secref{sec:which_divergence}, mode collapse does not seem to be
caused by any particular cost function.
It is commonly asserted that mode collapse is caused by the use of Jensen-Shannon
divergence, but this does not seem to be the case, because GANs that minimize
approximations of $\KL(\pdata \Vert \pmodel)$ face the same issues, and because
the generator often collapses to even fewer modes than would be preferred by the
Jensen-Shannon divergence.

\begin{figure}
\centering
\includegraphics[width=\textwidth]{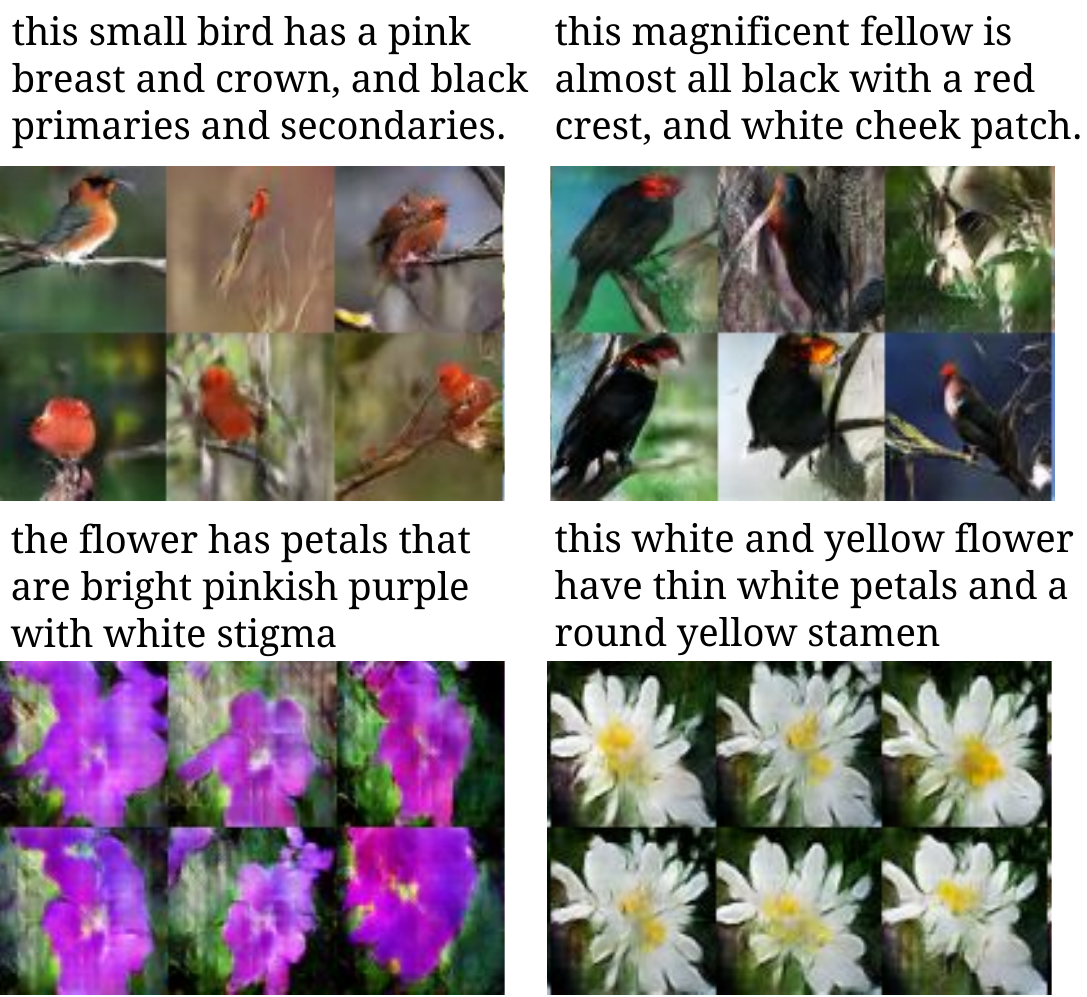}
\caption{
Text-to-image synthesis with GANs.
Image reproduced from \citet{reed2016generative}.
}
\label{fig:text2im}
\end{figure}

\begin{figure}
  \centering
  \includegraphics[width=\textwidth]{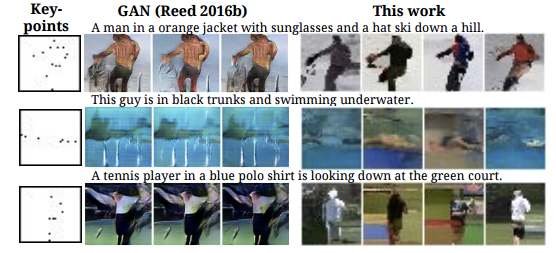}
  \caption{GANs have low output diversity for text-to-image
    tasks because of the mode collapse problem.
    Image reproduced from \citet{reedgenerating}.
  }
  \label{fig:low_diversity}
\end{figure}

\begin{figure}
  \centering
  \includegraphics[width=\textwidth]{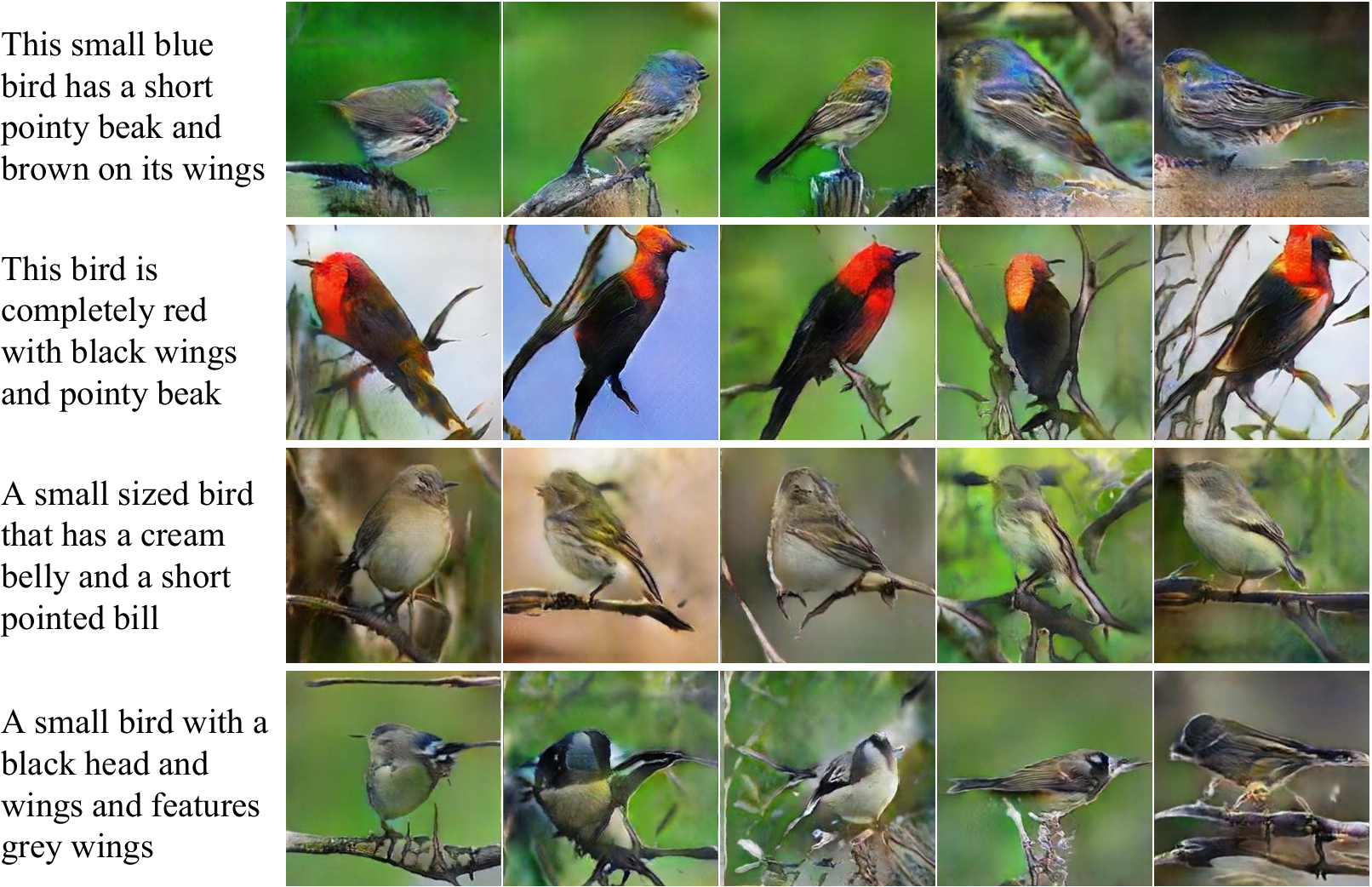}
  \caption{StackGANs are able to achieve higher output
    diversity than other GAN-based text-to-image models.
    Image reproduced from \citet{zhang2016stackgan}.}
    \label{fig:stackgan}
  \end{figure}

Because of the mode collapse problem, applications of GANs are often limited to
problems where it is acceptable for the model to produce a small number of 
distinct outputs, usually tasks where the goal is to map some input to one of
many acceptable outputs.
As long as the GAN is able to find a small number of these acceptable outputs,
it is useful.
One example is text-to-image synthesis, in which the input is a caption for an
image, and the output is an image matching that description.
See \figref{fig:text2im} for a demonstration of this task.
In very recent work, \citet{reedgenerating} have shown that other models have
higher output diversity than GANs for such tasks (\figref{fig:low_diversity}),
but StackGANs \citep{zhang2016stackgan} seem to have higher output diversity than previous GAN-based
approaches (\figref{fig:stackgan}).

The mode collapse problem is probably the most important issue with GANs that
researchers should attempt to address.

One attempt is \newterm{minibatch features} \citep{salimans2016improved}.
The basic idea of minibatch features is to allow the discriminator to compare
an example to a minibatch of generated samples and a minibatch of real samples.
By measuring distances to these other samples in latent spaces, the discriminator
can detect if a sample is unusually similar to other generated samples.
Minibatch features work well.
It is strongly recommended to directly copy the Theano/TensorFlow code released
with the paper that introduced them, since small changes in the definition of the
features result in large reductions in performance.

\begin{figure}
  \centering
  \includegraphics[width=\figwidth]{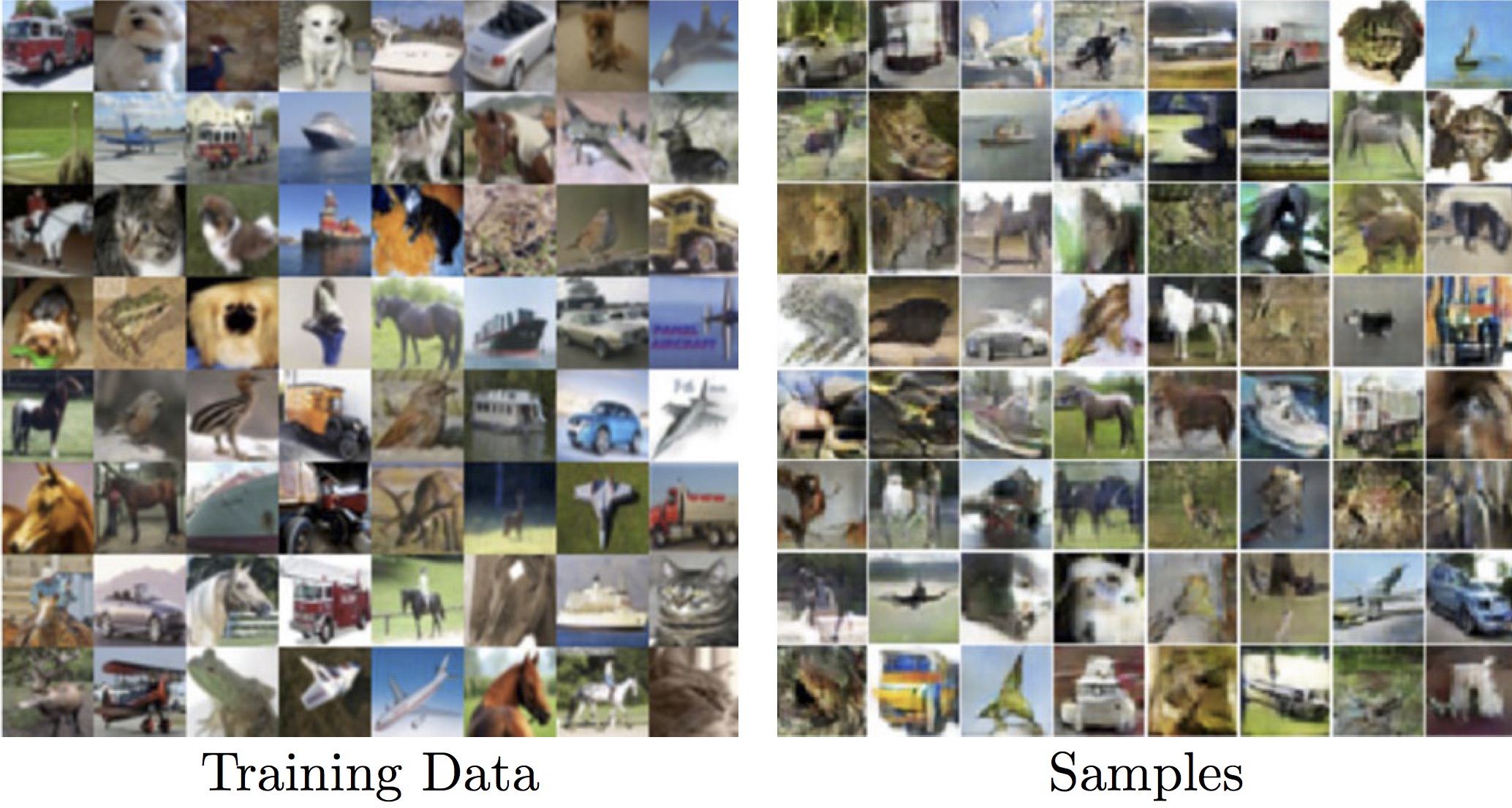}
  \caption{
    Minibatch GANs trained on CIFAR-10 obtain excellent results, with most samples
    being recognizable as specific CIFAR-10 classes.
    (Note: this model was trained with labels)
  }
  \label{fig:minibatch_cifar}
\end{figure}

\begin{figure}
  \centering
  \includegraphics[width=\figwidth]{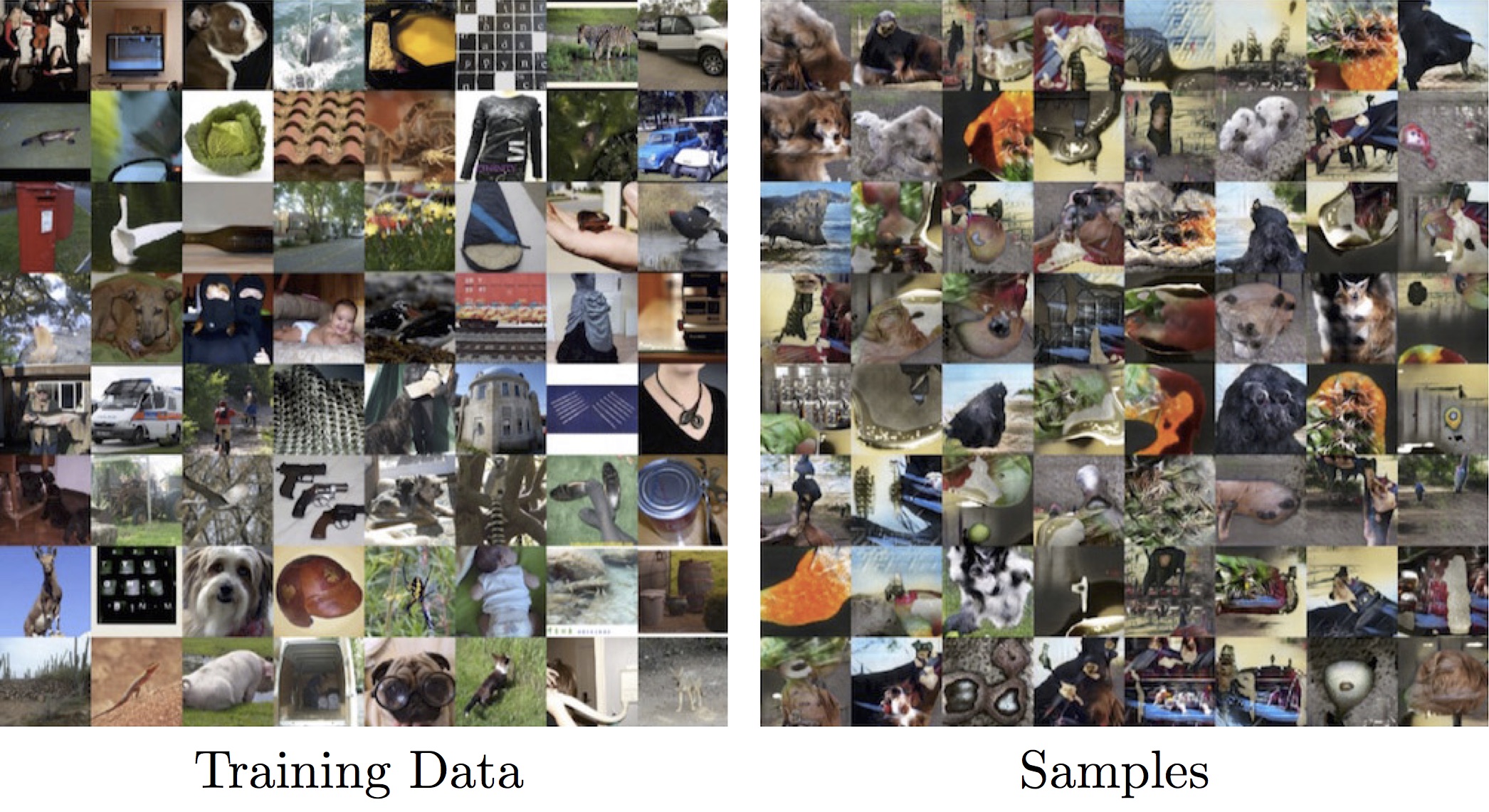}
  \caption{
    Minibatch GANS trained with labels on $128 \times 128$ ImageNet produce images
    that are occasionally recognizable as belonging to specific
    classes.
  }
  \label{fig:minibatch_imagenet}
\end{figure}

\begin{figure}
  \centering
  \includegraphics[width=\figwidth]{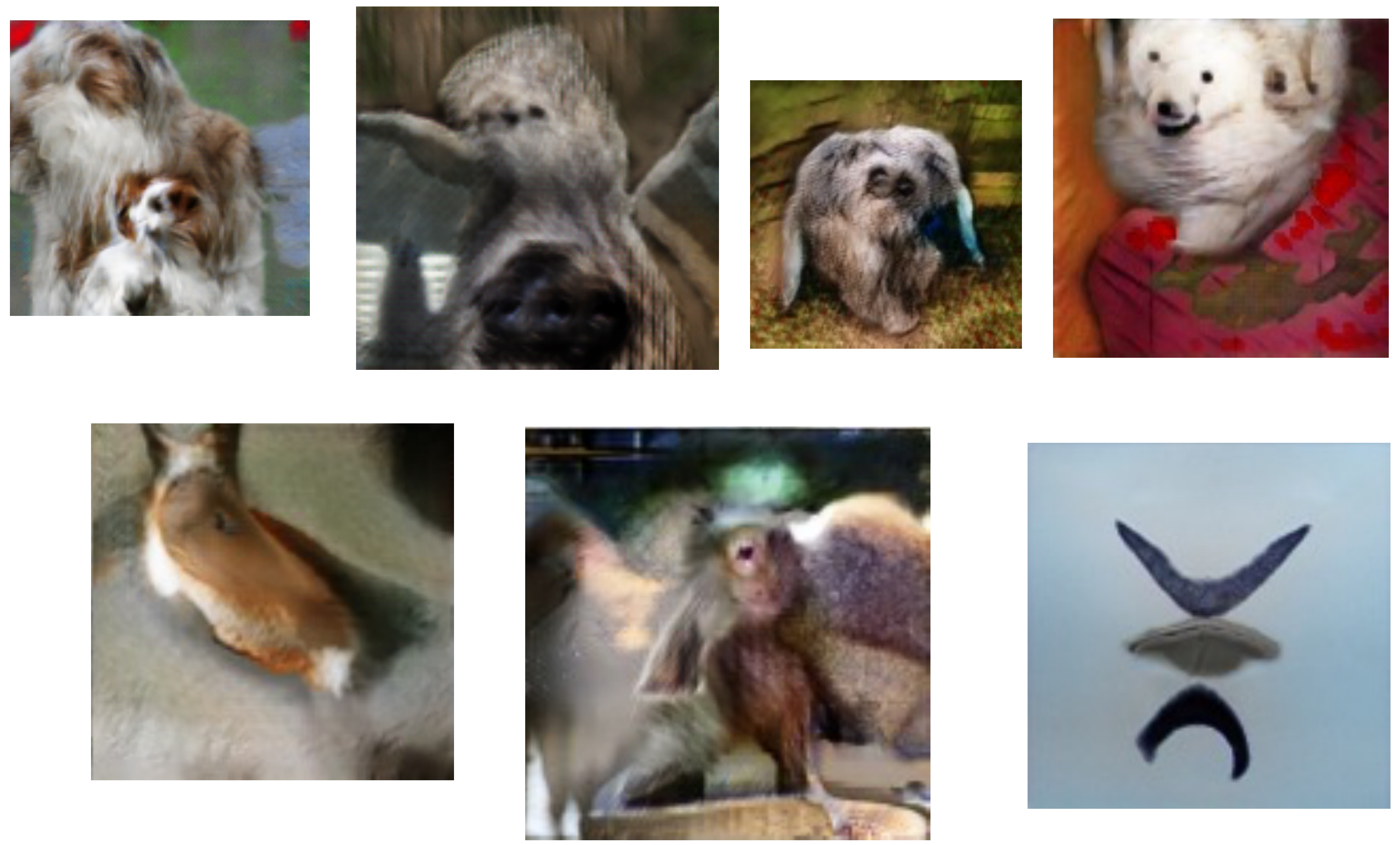}
  \caption{
    Minibatch GANs sometimes produce very good images when trained on $128 \times 128$
    ImageNet, as demonstrated by these cherry-picked examples.
  }
  \label{fig:cherry}
\end{figure}

Minibatch GANs trained on CIFAR-10 obtain excellent results, with most samples
being recognizable as specific CIFAR-10 classes (\figref{fig:minibatch_cifar}).
When trained on $128 \times 128$ ImageNet, few images are recognizable as belonging
to a specific ImageNet class (\figref{fig:minibatch_imagenet}).
Some of the better images are cherry-picked into \figref{fig:cherry}.

\begin{figure}
  \centering
  \includegraphics[width=\figwidth]{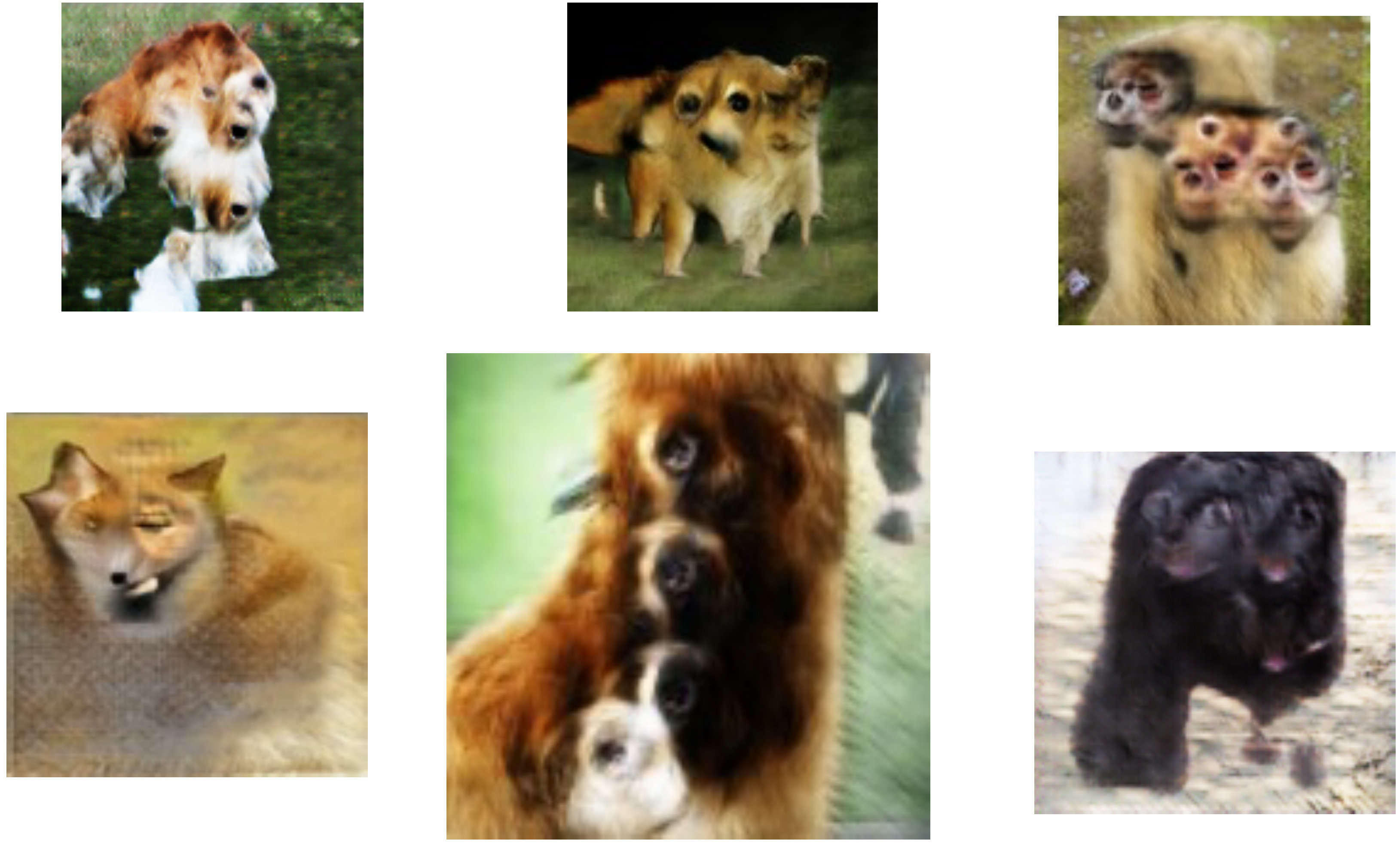}
  \caption{
    GANs on $128\times 128$ ImageNet seem to have trouble with counting, often generating
    animals with the wrong number of body parts.
  }
  \label{fig:counting}
\end{figure}

\begin{figure}
  \centering
  \includegraphics[width=\figwidth]{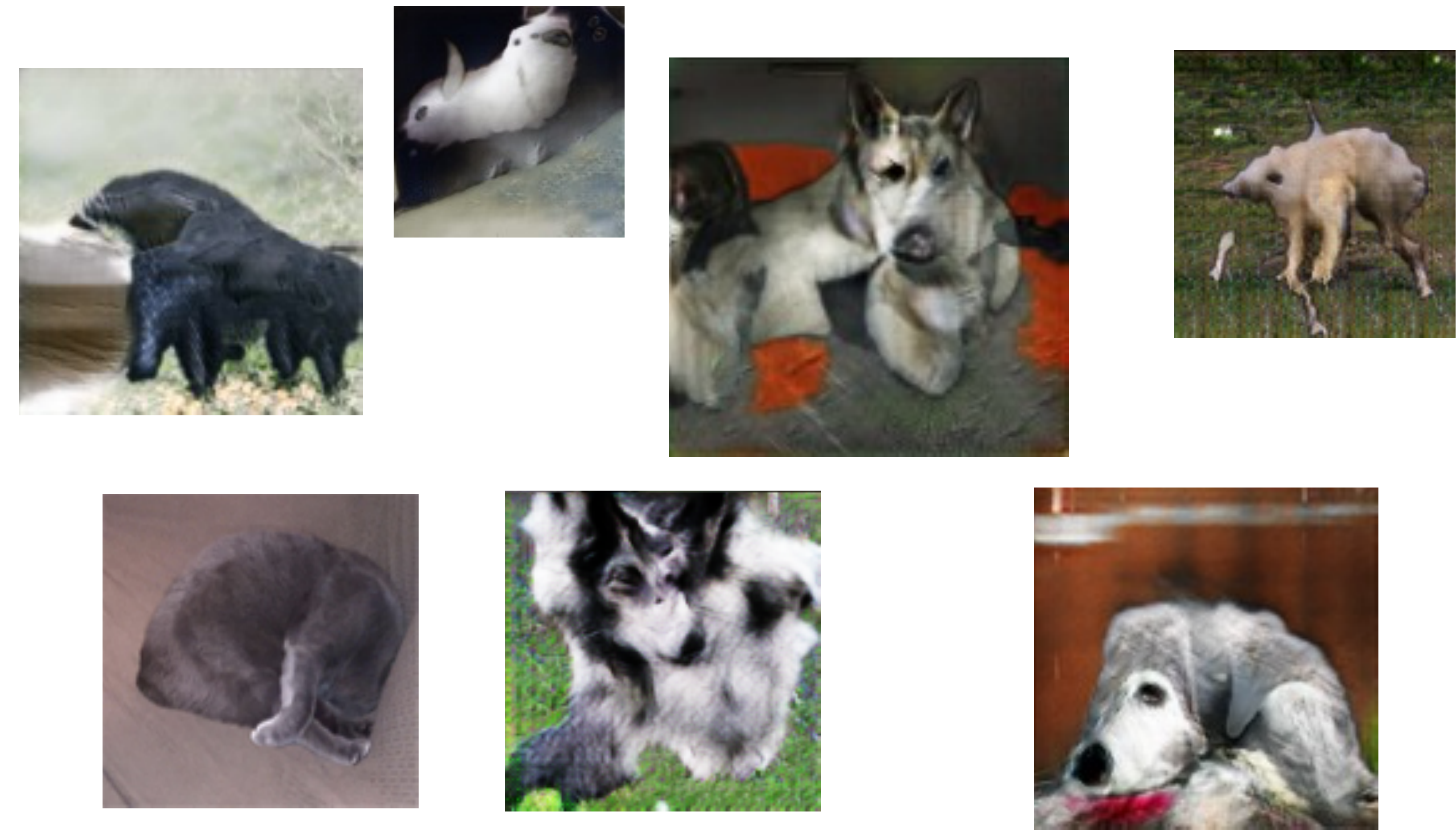}
  \caption{
    GANs on $128\times 128$ ImageNet seem to have trouble with the idea of three-dimensional
    perspective, often generating images of objects that are too flat or highly axis-aligned.
    As a test of the reader's discriminator network, one of these images is actually real.
  }
  \label{fig:perspective}
\end{figure}

\begin{figure}
  \centering
  \includegraphics[width=\figwidth]{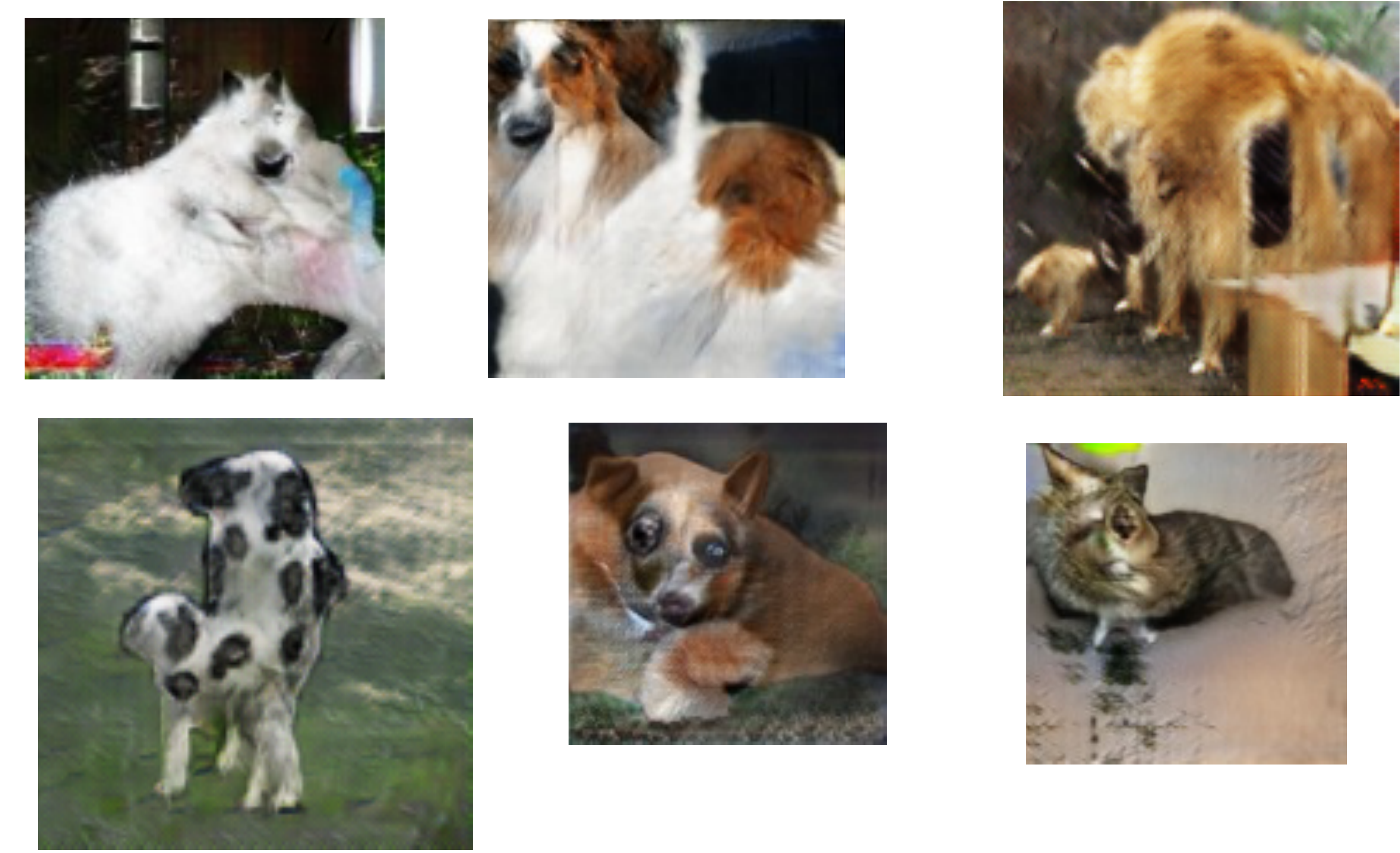}
  \caption{
    GANs on $128\times 128$ ImageNet seem to have trouble coordinating global structure,
    for example, drawing ``Fallout Cow,'' an animal that has both quadrupedal and bipedal structure.
  }
  \label{fig:structure}
\end{figure}

Minibatch GANs have reduced the mode collapse problem enough that other problems, such
as difficulties with counting, perspective, and global structure become the most obvious
defects (\figref{fig:counting}, \figref{fig:perspective}, and \figref{fig:structure},
respectively).
Many of these problems could presumably be resolved by designing better model architectures.

\begin{figure}
  \centering
  \includegraphics[width=\figwidth]{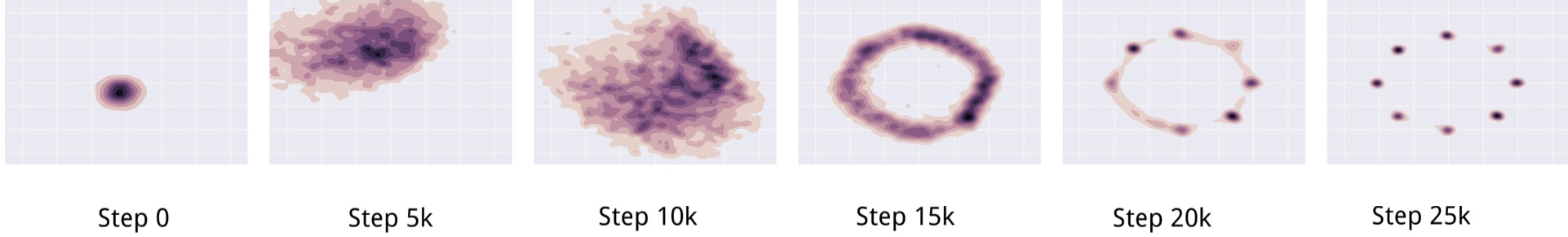}
  \caption{Unrolled GANs are able to fit all of the modes of a mixture of Gaussians
    in a two-dimensional space. Image reproduced from \citet{metz2016unrolled}.
  }
  \label{fig:unrolled}
\end{figure}

Another approach to solving the mode collapse problem is \newterm{unrolled GANs} \citep{metz2016unrolled}.
Ideally, we would like to find $G^* = \argmin_G \max_D V(G, D)$.
In practice, when we simultaneously follow the gradient of $V(G, D)$ for both players, we essentially ignore
the $\max$ operation when computing the gradient for $G$.
Really, we should regard $\max_D V(G,D)$ as the cost function for $G$, and we should back-propagate through
the maximization operation.
Various strategies exist for back-propagating through a maximization operation, but many, such as those
based on implicit differentiation, are unstable.
The idea of unrolled GANs is to build a computational graph describing $k$ steps of learning
in the discriminator, then backpropagate through all $k$ of these steps of learning
when computing the gradient on the generator.
Fully maximizing the value function for the discriminator takes tens of thousands of steps,
but \citet{metz2016unrolled} found that unrolling for even small numbers of steps, like 10 or fewer,
can noticeably reduce the mode dropping problem.
This approach has not yet been scaled up to ImageNet.
See \figref{fig:unrolled} for a demonstration of unrolled GANs on a toy problem.

\subsubsection{Other games}

If our theory of how to understand whether a continuous, high-dimensional non-convex game 
will converge could be improved, or if we could develop algorithms that converge 
more reliably than simultaneous gradient descent, several application areas besides GANs
would benefit.
Even restricted to just AI research, we find games in many scenarios:
\begin{itemize}
  \item Agents that literally play games, such as AlphaGo \citep{silver2016mastering}.
  \item Machine learning security, where models must resist adversarial examples \citep{Szegedy-ICLR2014,Goodfellow-2015-adversarial}.
  \item Domain adaptation via domain-adversarial learning \citep{ganin2015domain}.
  \item Adversarial mechanisms for preserving privacy \citep{edwards2015censoring}.
  \item Adversarial mechanisms for cryptography \citep{abadi2016learning}.
\end{itemize}
This is by no means an exhaustive list.

\subsection{Evaluation of generative models}

Another highly important research area related to GANs is that it is not clear
how to quantitatively evaluate generative models.
Models that obtain good likelihood can generate bad samples, and models that
generate good samples can have poor likelihood.
There is no clearly justified way to quantitatively score samples.
GANs are somewhat harder to evaluate than other generative models because
it can be difficult to estimate the likelihood for GANs
(but it is possible---see \citet{wu2016quantitative}).
\citet{Theis2015d} describe many of the difficulties with evaluating generative models.

\subsection{Discrete outputs}

The only real requirement imposed on the design of the generator by the GAN framework
is that the generator must be differentiable.
Unfortunately, this means that the generator cannot produce discrete data, such
as one-hot word or character representations.
Removing this limitation is an important research direction that could unlock the
potential of GANs for NLP.
There are at least three obvious ways one could attack this problem:
\begin{enumerate}
  \item Using the REINFORCE algorithm \citep{Williams-1992}.
  \item Using the concrete distribution \citep{maddison2016concrete} or Gumbel-softmax \citep{jang2016categorical}.
  \item Training the generate to sample continuous values that can be decoded to discrete ones (e.g., sampling
    word embeddings directly).
\end{enumerate}

\subsection{Semi-supervised learning}
\label{sec:ssl}

A research area where GANs are already highly successful is the use of generative
models for semi-supervised learning, as proposed but not demonstrated in the original
GAN paper \citep{Goodfellow-et-al-NIPS2014-small}.

GANs have been successfully applied to semi-supervised learning at least since the introduction
of CatGANs \citep{springenberg2015unsupervised}.
Currently, the state of the art in semi-supervised learning on MNIST, SVHN, and CIFAR-10
is obtained by \newterm{feature matching GANs} \citep{salimans2016improved}.
Typically, models are trained on these datasets using 50,000 or more labels,
but feature matching GANs are able to obtain good performance
with very few labels.
They obtain state of the art performance within several categories for different
amounts of labels, ranging from 20 to 8,000.

The basic idea of how to do semi-supervised learning with feature matching GANs
is to turn a classification problem with $n$ classes into a classification problem
with $n+1$ classes, with the additional class corresponding to fake images.
All of the real classes can be summed together to obtain the probability of the
image being real, enabling the use of the classifier as a discriminator within
 the GAN game.
 The real-vs-fake discriminator can be trained even with unlabeled data, which
 is known to be real, and with samples from the generator, which are known to
 be fake.
 The classifier can also be trained to recognize individual real classes on the limited
 amount of real, labeled examples.
 This approach was simultaneously developed by \citet{salimans2016improved}
 and \citet{odena2016semi}. The earlier CatGAN used an $n$ class discriminator
 rather than an $n+1$ class discriminator.

 Future improvements to GANs can presumably be expected to yield further
 improvements to semi-supervised learning.

 \subsection{Using the code}

 GANs learn a representation $\vz$ of the image $\vx$.
 It is already known that this representation can capture useful high-level
 abstract semantic properties of $\vx$, but it can be somewhat difficult
 to make use of this information.

One obstacle to using $\vz$ is that it can be difficult to obtain
$\vz$ given an input $\vx$.
\citet{Goodfellow-et-al-NIPS2014-small} proposed but did not demonstrate
using a second network analogous to the generator to sample from $p(\vz \mid \vx)$,
much as the generator samples from $p(\vx)$.
So far the full version of this idea, using a fully general neural network as the
encoder and sampling from an arbitrarily powerful approximation of $p(\vz \mid \vx)$,
has not been successfully demonstrated,
but
\citet{donahue2016adversarial} demonstrated how to train a deterministic encoder,
and \citet{dumoulin2016adversarially} demonstrated how to train an encoder
network that samples from a Gaussian approximation of the posterior.
Futher research will presumably develop more powerful stochastic encoders.

Another way to make better use of the code is to train the code to be more useful.
InfoGANs \citep{chen2016infogan} regularize some entries in the code vector with
an extra objective function that encourages them to have high mutual information
with $\vx$. Individual entries in the resulting code then correspond to specific
semantic attributes of $\vx$, such as the direction of lighting on an image of
a face.

\subsection{Developing connections to reinforcement learning}
\label{sec:rl_connections}

Researchers have already identified connections between GANs and
actor-critic methods \citep{pfau2016connecting}, inverse reinforcement learning \citep{finn2016connection},
and have applied GANs to imitation learning \citep{ho2016generative}.
These connections to RL will presumably continue to bear fruit, both for GANs
and for RL.

\section{Plug and Play Generative Networks}

Shortly before this tutorial was presented at NIPS, a new generative model
was released. This model, plug and play generative networks \citep{nguyen2016plug}, has dramatically
improved the diversity of samples of images of ImageNet classes that can be
produced at high resolution.

PPGNs are new and not yet well understood.
The model is complicated, and most of the recommendations about how to design the model
are based on empirical observation rather than theoretical understanding.
This tutorial will thus not say too much about exactly how PPGNs work, since this will
presumably become more clear in the future.

As a brief summary, PPGNs are basically an approximate Langevin sampling approach to
generating images with a Markov chain.
The gradients for the Langevin sampler are estimated using a denoising autoencoder.
The denoising autoencoder is trained with several losses, including a GAN loss.

\begin{figure}
\centering
\includegraphics[width=\figwidth]{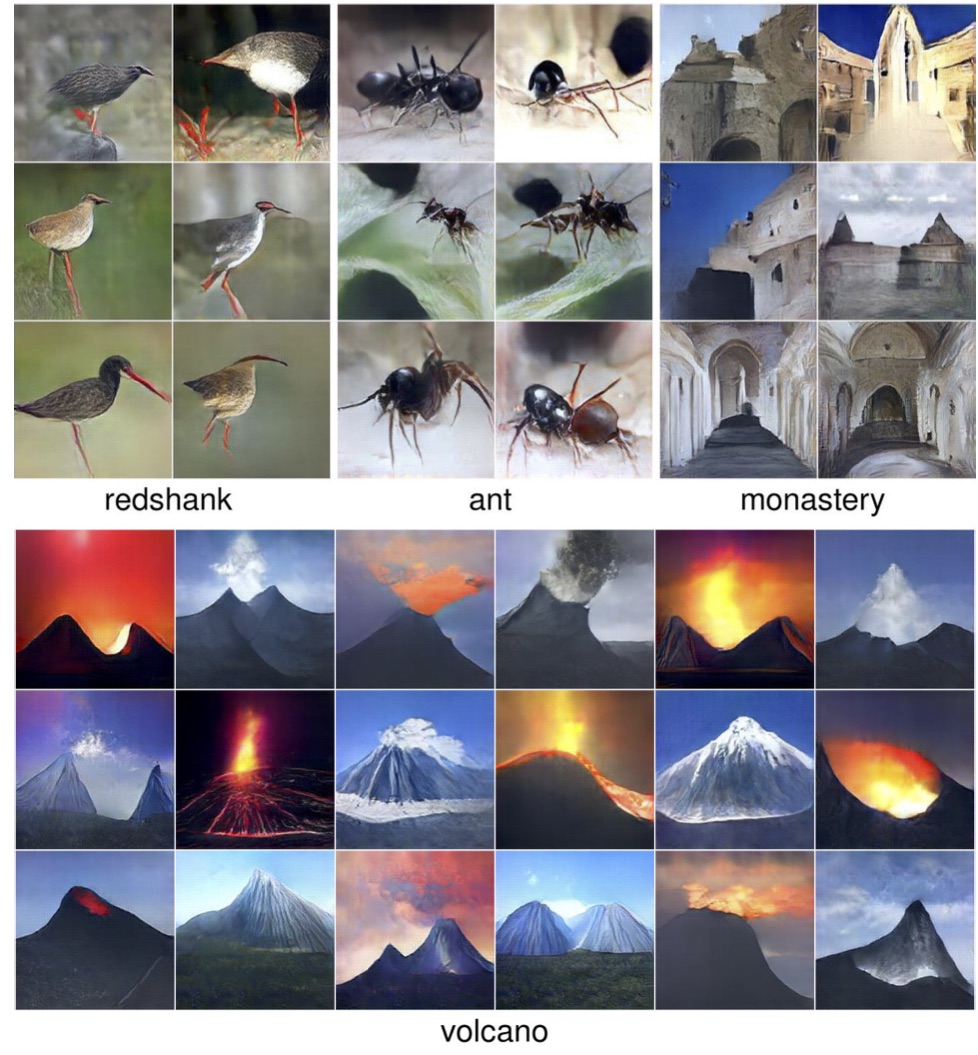}
\caption{PPGNs are able to generate diverse, high resolution images from ImageNet
classes. Image reproduced from \citet{nguyen2016plug}.}
\label{fig:ppgn}
\end{figure}

\begin{figure}
\centering
\includegraphics[width=\figwidth]{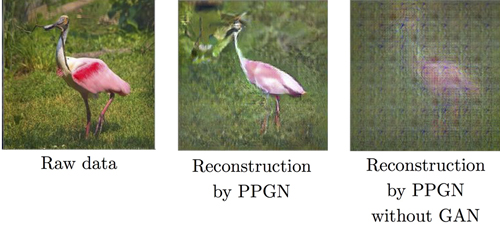}
\caption{The GAN loss is a crucial ingredient of PPGNs. Without it, the denoising autoencoder
used to drive PPGNs does not create compelling images.}
\label{fig:recons}
\end{figure}

Some of the results are shown in \figref{fig:ppgn}.
As demonstrated in \figref{fig:recons}, the GAN loss is crucial for obtaining high quality images.

\section{Exercises}

This tutorial includes three exercises to check your understanding.
The solutions are given in \secref{sec:solutions}.

\subsection{The optimal discriminator strategy}
\label{sec:opt_d}

As described in \eqref{eq:discriminator_cost}, the goal of the discriminator is to minimize
\begin{equation}
  J^{(D)}(\vtheta^{(D)}, \vtheta^{(G)}) = -\frac{1}{2} \E_{\vx \sim \pdata} \log D(\vx) - \frac{1}{2} \E_{\vz} \log \left(1 - D\left( G(z) \right) \right)
\end{equation}
with respect to $\vtheta^{(D)}$.
Imagine that the discriminator can be optimized in function space, so the value of
$D(\vx)$ is specified independently for every value of $\vx$.
What is the optimal strategy for $D$?
What assumptions need to be made to obtain this result?

\subsection{Gradient descent for games}
\label{sec:xy_exercise}

Consider a minimax game with two players that each control a single scalar value.
The minimizing player controls scalar $x$ and the maximizing player controls
scalar $y$.
The value function for this game is
\[ V(x, y) = x y .\]

\begin{itemize}
  \item Does this game have an equilibrium? If so, where is it?
  \item Consider the learning dynamics of simultaneous gradient descent.
        To simplify the problem, treat gradient descent as a continuous time
        process.
        With an infinitesimal learning rate, gradient descent is described by a system of partial differential equations:
        \begin{align}
          \frac{\partial x}{\partial t} &= - \frac{\partial}{\partial x} V\left( x(t), y(t) \right) \\
          \frac{\partial y}{\partial t} &= \frac{\partial}{\partial y} V\left( x(t), y(t) \right).
        \end{align}
        Solve for the trajectory followed by these dynamics.
\end{itemize}

\subsection{Maximum likelihood in the GAN framework}
\label{sec:mle_exercise}

In this exercise, we will derive a cost that yields (approximate) maximum likelihood learning within the GAN framework.
Our goal is to design $J^{(G)}$ so that, if we assume the discriminator is optimal, the expected gradient of $J^{(G)}$
will match the expected gradient of $\KL(\pdata \Vert \pmodel)$.

The solution will take the form of:
\[
  J^{(G)} = \E_{\vx \sim p_g} f(\vx) .
\]

The exercise consists of determining the form of $f$.

\section{Solutions to exercises}
\label{sec:solutions}

\subsection{The optimal discriminator strategy}
\label{sec:opt_d_soln}

Our goal is to minimize
\begin{equation}
  J^{(D)}(\vtheta^{(D)}, \vtheta^{(G)}) = -\frac{1}{2} \E_{\vx \sim \pdata} \log D(\vx) - \frac{1}{2} \E_{\vz} \log \left(1 - D\left( G(z) \right) \right)
\end{equation}
in function space, specifying $D(\vx)$ directly.

We begin by assuming that both $\pdata$ and $\pmodel$ are nonzero everywhere.
If we do not make this assumption, then some points are never visited during training,
and have undefined behavior.

To minimize $J^{(D)}$ with respect to $D$, we can write down the functional derivatives with
respect to a single entry $D(\vx)$, and set them equal to zero:
\[
\frac{\delta} {\delta D(\vx)} J^{(D)} = 0.
\]
By solving this equation, we obtain
\[
D^*(\vx) = \frac{ \pdata(\vx) } {\pdata(\vx) + \pmodel(\vx) }.
\]

Estimating this ratio is the key approximation mechanism used by GANs.

\begin{figure}
\centering
\includegraphics[width=\figwidth]{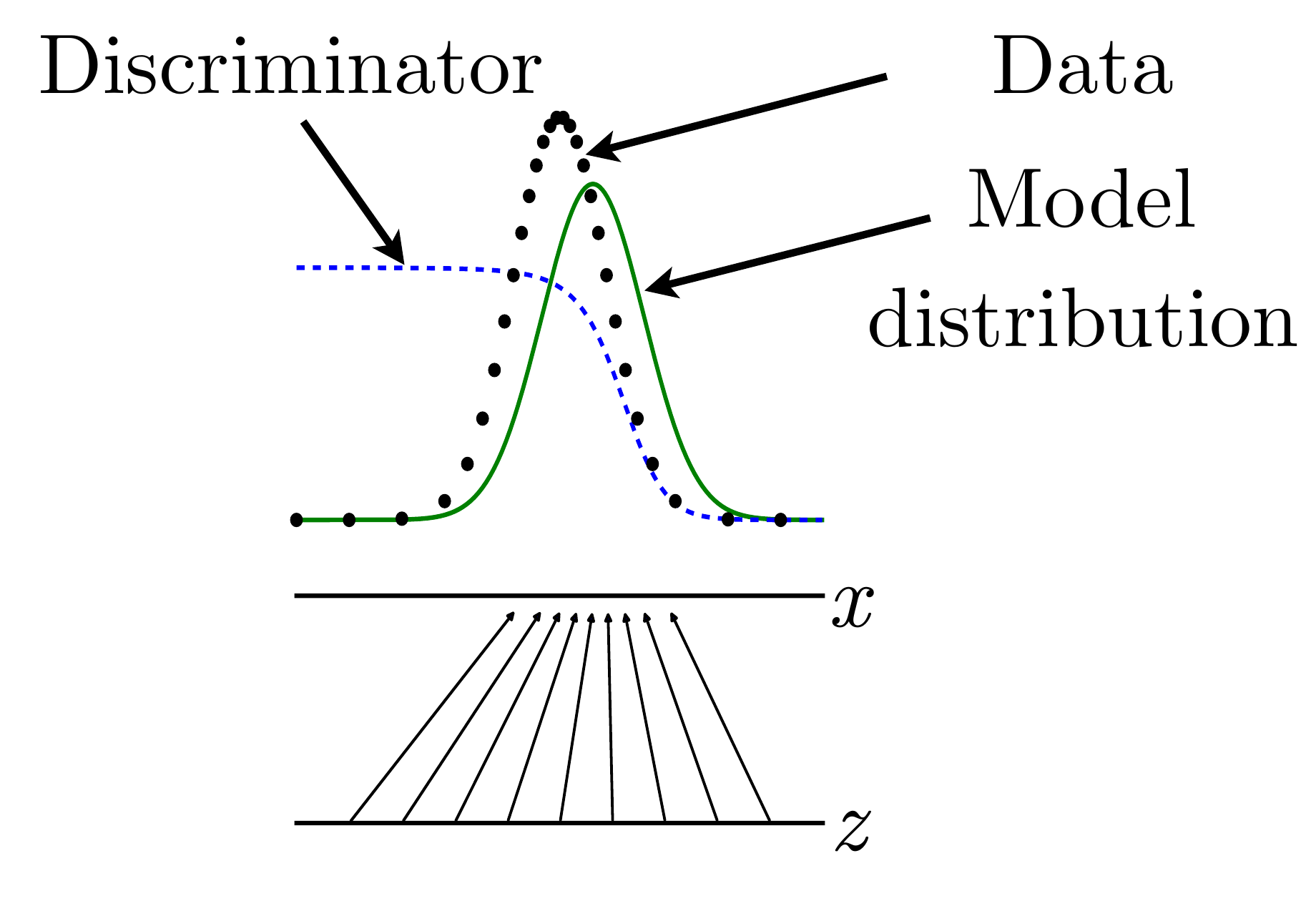}
\caption{
An illustration of how the discriminator estimates a ratio of
densities.
In this example, we assume that both $z$ and $x$ are one dimensional
for simplicity.
The mapping from $z$ to $x$ (shown by the black arrows) is non-uniform so that $\pmodel(x)$
(shown by the green curve) is
greater in places where $z$ values are brought together more densely.
The discriminator (dashed blue line) estimates the ratio between the data density (black dots)
and the sum of the data and model densities.
Wherever the output of the discriminator is large, the model density is too low, and wherever
the output of the discriminator is small, the model density is too high.
The generator can learn to produce a better model density by following the discriminator uphill;
each $G(z)$ value should move slightly in the direction that increases $D(G(z))$.
Figure reproduced from \citet{Goodfellow-et-al-NIPS2014-small}.
}
\label{fig:ratio}
\end{figure}

The process is illustrated in \figref{fig:ratio}.

\subsection{Gradient descent for games}
\label{sec:xy_soln}

\begin{figure}
\centering
\includegraphics[width=\figwidth]{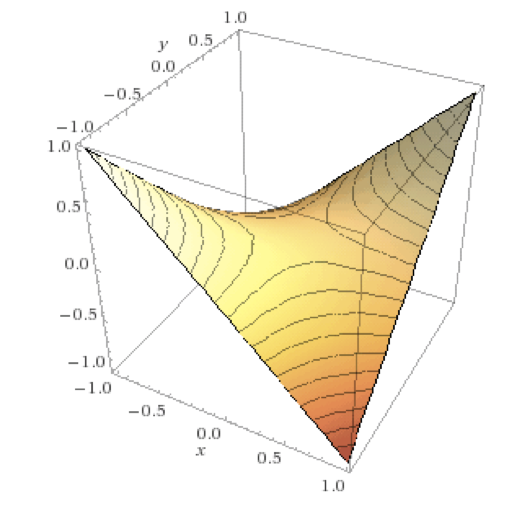}
\caption{A three-dimensional visualization of the value function $V(x,y) = xy$.
  This is the canonical example of a function with a saddle point, at $x=y=0$.
}
\label{fig:xy}
\end{figure}

The value function
\[ V(x, y) = x y \]
is the simplest possible example of a continuous function with a saddle point.
It is easiest to understand this game by visualizing the value function in three
dimensions, as shown in \figref{fig:xy}.

The three dimensional visualization shows us clearly that there is a saddle point
at $x=y=0$. This is an equilibrium of the game. We could also have found this point
by solving for where the derivatives are zero.

Not every saddle point is an equilibrium; we require that an infinitesimal perturbation
of one player's parameters cannot reduce that player's cost.
The saddle point for this game satisfies that requirement.
It is something of a pathological equilibrium because the value function is constant
as a function of each player's parameter when holding the other player's parameter
fixed.

To solve for the trajectory taken by gradient descent, we take the derivatives, and find that
\begin{align}
  \frac{\partial x}{\partial t} = - y(t) \\
  \frac{\partial y}{\partial t} = x(t). \label{eq:dy}
\end{align}
Diffentiating \eqref{eq:dy}, we obtain
\[
  \frac{\partial^2 y}{\partial t^2} = \frac{\partial x}{\partial t} = -y(t).
\]
Differential equations of this form have sinusoids as their set of basis functions
of solutions.
Solving for the coefficients that respect the boundary conditions, we obtain
\begin{align}
  x(t) = x(0) \cos(t) - y(0) \sin(t) \\
  y(t) = x(0) \sin(t) + y(0) \cos(t).
\end{align}

\begin{figure}
  \center
  \includegraphics[width=\figwidth]{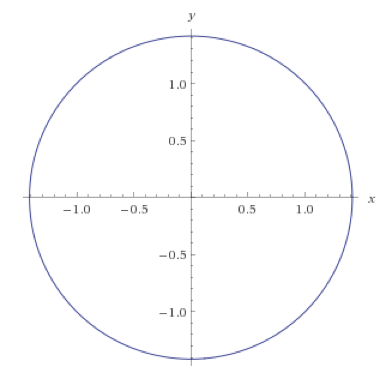}
  \caption{Simultaneous gradient descent with infinitesimal learning rate
    will orbit indefinitely at constant radius when applied to $V(x,y) = xy$,
    rather than approaching the equilibrium solution at $x=y=0$.
  }
  \label{fig:orbit}
\end{figure}

These dynamics form a circular orbit, as shown in \figref{fig:orbit}.
In other words, simultaneous gradient descent with an infinitesimal learning rate
will orbit the equilibrium forever, at the same radius that it was initialized.
With a larger learning rate, it is possible for simultaneous gradient descent to
spiral outward forever.
Simultaneous gradient descent will never approach the equilibrium.

For some games, simultaneous gradient descent does converge, and for others,
such as the one in this exercise, it does not.
For GANs, there is no theoretical prediction as to whether simultaneous
gradient descent should converge or not.
Settling this theoretical question, and developing algorithms guaranteed to
converge, remain important open research problems.

\subsection{Maximum likelihood in the GAN framework}
\label{sec:mle_soln}

We wish to find a function $f$ such that the expected gradient of 
\begin{equation}
  J^{(G)} = \E_{\vx \sim p_g} f(\vx)
  \label{eq:cost_per_sample}
\end{equation}
is equal to the expected gradient of 
$\KL(\pdata \Vert p_g)$.

First we take the derivative of the KL divergence with respect to a parameter $\theta$:
\begin{equation}
  \frac{\partial}{\partial \theta} \KL(\pdata \Vert p_g) = - \E_{\vx \sim \pdata} \frac{\partial}{\partial \theta} \log p_g(\vx) .
\label{eq:mle_gradient}
\end{equation}

We now want to find the $f$ that will make the derivatives of \eqref{eq:cost_per_sample} match \eqref{eq:mle_gradient}.
We begin by taking the derivatives of \eqref{eq:cost_per_sample}:
\[
  \frac{\partial}{\partial \theta} J^{(G)} = \E_{\vx \sim p_g} f(x) \frac{\partial}{\partial \theta} \log p_g(\vx).
\]
To obtain this result, we made two assumptions:
\begin{enumerate}
  \item We assumed that $p_g(\vx) \geq 0$ everywhere so that we were able to use the identity $p_g(\vx) = \exp( \log p_g(\vx) ).$
  \item We assumed that we can use Leibniz's rule to exhange the order of differentiation and integration (specifically, that both the function and its derivative are continuous, and that the function vanishes for infinite values of $\vx$).
\end{enumerate}

We see that the derivatives of $J^{(G)}$ come very near to giving us what we want; the only problem is that
the expectation is computed by drawing samples from $p_g$ when we would like it to be computed by drawing
samples from $\pdata$.
We can fix this problem using an importance sampling trick; by setting $f(x) = \frac{\pdata(\vx)}{p_g(\vx)}$
we can reweight the contribution to the gradient from each generator sample to compensate for it having
been drawn from the generator rather than the data.

Note that when constructing $J^{(G)}$ we must {\em copy} $p_g$ into $f(x)$ so that $f(x)$ has a derivative of
zero with respect to the parameters of $p_g$.
Fortunately, this happens naturally if we obtain the value of $\frac{\pdata(\vx)}{p_g(\vx)}$.

From \secref{sec:opt_d_soln}, we already know that the discriminator estimates the desired ratio.
Using some algebra, we can obtain a numerically stable implementation of $f(\vx)$.
If the discriminator is defined to apply a logistic sigmoid function at the output layer,
with $D(\vx) = \sigma( a(\vx) )$, then $f(x) = - \exp(a(\vx))$.

This exercise is taken from a result shown by \citet{Goodfellow-ICLR2015}.
From this exercise, we see that the discriminator estimates a ratio of densities
that can be used to calculate a variety of divergences.

\section{Conclusion}

GANs are generative models that use supervised learning to approximate an intractable cost
function, much as Boltzmann machines use Markov chains to approximate their cost and VAEs
use the variational lower bound to approximate their cost.
GANs can use this supervised ratio estimation technique to approximate many cost functions, including the KL divergence used for maximum
likelihood estimation.

GANs are relatively new and still require some research to reach their new potential.
In particular, training GANs requires finding Nash equilibria in high-dimensional,
continuous, non-convex games.
Researchers should strive to develop better theoretical understanding and better training
algorithms for this scenario.
Success on this front would improve many other applications, besides GANs.

GANs are crucial to many different state of the art image generation and manipulation systems,
and have the potential to enable many other applications in the future.

\section*{Acknowledgments}
The author would like to thank the NIPS organizers for inviting him to
present this tutorial.
Many thanks also to those who commented on his Twitter and Facebook posts
asking which topics would be of interest to the tutorial audience.
Thanks also to D. Kingma for helpful discussions regarding the description of VAEs.
Thanks to Zhu Xiaohu, Alex Kurakin and Ilya Edrenkin for spotting typographical errors in the
manuscript.

\bibliography{biblio}
\bibliographystyle{natbib}

\end{document}